%% file: main.tex
\pdfoutput=1

\documentclass[11pt]{article}

\usepackage[]{acl}

\usepackage{times}
\usepackage{latexsym}

\usepackage[T1]{fontenc}

\usepackage[utf8]{inputenc}

\usepackage{microtype}

\usepackage{inconsolata}

\usepackage{graphicx}

\usepackage{xspace}
\usepackage{amsmath}
\usepackage{booktabs}
\usepackage{tcolorbox}
\usepackage{listings}
\usepackage{multirow}
\usepackage{soul}
\usepackage{bm}
\usepackage{relsize}
\usepackage{longtable}

\newcommand{\DarkRed}[1]{\textcolor[rgb]{0.75,0.00,0.00}{#1}}

\newcommand{\revise}[1]{{#1}}

\newcommand{\ourdataset}{\textsc{AoT Collection}\xspace}
\newcommand{\aottocot}{AoT2CoT\xspace}
\newcommand{\ourcot}{CoT (Ablation)\xspace}

\definecolor{error}{HTML}{F47874}
\DeclareRobustCommand{\hlerror}[1]{{\sethlcolor{error}\hl{#1}}}
\title{Abstraction-of-Thought Makes Language Models Better Reasoners}

\author{Ruixin Hong\textsuperscript{1,2}\thanks{~~Work done during the internship at Tencent AI Lab.}, 
Hongming Zhang\textsuperscript{2}, 
Xiaoman Pan\textsuperscript{2}, 
Dong Yu\textsuperscript{2}, 
Changshui Zhang\textsuperscript{1} \\
\textsuperscript{1}Institute for Artificial Intelligence, Tsinghua University (THUAI); \\
\textsuperscript{1}Beijing National Research Center for Information Science and Technology (BNRist); \\
\textsuperscript{1}Department of Automation, Tsinghua University, Beijing, P.R.China \\
\textsuperscript{2}Tencent AI Lab, Seattle \\
\texttt{hrx20@mails.tsinghua.edu.cn,} 
\texttt{\{hongmzhang, dyu\}@tencent.com,} \\
\texttt{shyowmanpan@gmail.com,} \texttt{zcs@mail.tsinghua.edu.cn,}
}

\begin{document}
\maketitle
\begin{abstract}

Abstract reasoning, the ability to reason from the abstract essence of a problem, serves as a key to generalization in human reasoning. 
However, eliciting language models to perform reasoning with abstraction remains unexplored.
This paper seeks to bridge this gap by introducing a novel structured reasoning format called Abstraction-of-Thought (AoT).
The uniqueness of AoT lies in its explicit requirement for varying levels of abstraction within the reasoning process. 
This approach could elicit language models to first contemplate on the abstract level before incorporating concrete details, 
which is overlooked by the prevailing step-by-step Chain-of-Thought (CoT) method.
To align models with the AoT format, we present \ourdataset, a generic finetuning dataset consisting of 348k high-quality samples with AoT reasoning processes, collected via an automated and scalable pipeline.
We finetune a wide range of language models with \ourdataset and conduct extensive evaluations on 23 unseen tasks from the challenging benchmark Big-Bench Hard.
Experimental results indicate that models aligned to AoT reasoning format substantially outperform those aligned to CoT in many reasoning tasks.\footnote{~~Data, code and models are available at \url{https://github.com/Raising-hrx/Abstraction-of-Thought}}.
\end{abstract}

\section{Introduction}

The complex reasoning ability is one of the long-term pursuits of artificial intelligence. 
In recent years, language models (LMs) have seen rapid development and achieved impressive performance on a variety of reasoning benchmarks~\cite{DBLP:conf/nips/BrownMRSKDNSSAA20,DBLP:journals/corr/abs-2303-08774}.
Among the advancements in reasoning methods, the Chain-of-Thought (CoT) technique has emerged as a prominent reasoning tool~\cite{DBLP:conf/nips/Wei0SBIXCLZ22}.
This technique, when employed in large language models (LLMs), serves as a guide that enables the model to initially generate intermediate reasoning processes before ultimately producing the final answer.
The generated reasoning processes
significantly help large language models improve their reasoning performance in zero-shot and few-shot scenarios. 
This discovery has sparked a great deal of research interest and further improvement~\cite{DBLP:journals/corr/abs-2309-15402,DBLP:journals/corr/abs-2401-14295,DBLP:conf/nips/KojimaGRMI22,DBLP:conf/iclr/ZhouSHWS0SCBLC23,DBLP:conf/nips/YaoYZS00N23}.
Meanwhile, for medium-scale language models, training models on data containing CoT has been proven to be an effective method to enhance the reasoning performance of language models~\cite{DBLP:journals/corr/abs-2402-13116,DBLP:conf/nips/ZelikmanWMG22,DBLP:conf/acl/ShridharSS23,DBLP:conf/acl/HoSY23,DBLP:conf/icml/FuPOSK23,DBLP:journals/corr/abs-2307-02053}.

\begin{figure}[t]
    \centering
    \includegraphics[width=\columnwidth]{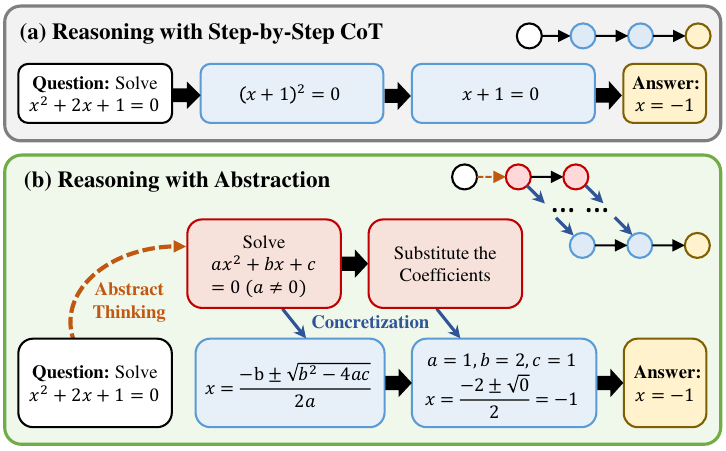}
    \caption{Reasoning with abstraction attempts to answer questions from the perspective of abstract essences, which may be overlooked by step-by-step Chain-of-Thought (CoT) reasoning. 
    The reasoning process with abstraction contains multiple levels of abstract information.
    The lower levels (blue nodes) are responsible for performing concrete reasoning and are typically rich in detail.
    Conversely, the higher levels (red nodes) are abstractions of lower levels, organizing the entire reasoning process.
    }
    \label{fig:intro}
\end{figure}

The prevailing CoT paradigm typically follows a sequential, step-by-step reasoning process, which may leave the exploration of abstraction in reasoning largely untapped.
Abstraction is the key to human cognition and reasoning~\cite{saitta2013abstraction}. 
When dealing with complex tasks, reasoning with abstraction first tackles the essence and the most crucial parts, then utilizes the abstract skeletal solution as a guide to solve the rest of the problem~\cite{yang2012intelligent}. 
Abstraction allows reasoning to perform at a higher and more essential level, resulting in a more generalizable and efficient solution.
\revise{As shown in Figure~\ref{fig:intro}, when dealing with a specific problem of solving a quadratic equation, reasoning with abstraction suggests first deriving the general quadratic formula and then substituting the specific coefficients to get the answer. 
This approach offers a more fundamental solution, applicable to problems with similar reasoning patterns. 
The existing CoT, in the absence of explicit constraints, does not guarantee that abstraction will necessarily be incorporated into the problem-solving process.}

In this paper, we explore how to elicit language models to reason with abstraction. 
\revise{We propose a novel format of reasoning processes named Abstraction-of-Thought (AoT), which is a constrained variant of the CoT.}
Reasoning with abstraction typically starts with a rough skeletal solution from an abstraction perspective (e.g., the red nodes in Figure~\ref{fig:intro}(b)), and then gradually concretizes the solution until the problem is solved.
We suggest that employing an abstract skeletal solution to organize the entire reasoning process could be the key to eliciting abstract reasoning. 
Thus, in contrast to the unconstrained CoT, the AoT format explicitly demands the inclusion of varying levels of abstraction throughout the reasoning process.
The higher level is an abstraction of the lower level, containing fewer concrete details but stating the objective and functionality of the reasoning step.
For example, functions and classes in programs are abstractions of the following specific code fragments, while main claims in argumentative discourses~\cite{DBLP:journals/coling/Cohen87} are abstractions of subsequent supportive evidence.
\revise{Specifically, we focus on the two-level AoT in this paper, which contains an additional level of abstraction over the CoT, to serve as an preliminary exploration of whether abstraction helps language models reasoning.}

To align language models to the AoT format, we present the \ourdataset, a supervised finetuning dataset that augments 348k AoT reasoning processes from the FLAN Collection~\cite{DBLP:conf/icml/LongpreHVWCTZLZ23}.
\ourdataset covers 216 generic tasks that are not specifically designed for a certain domain or dataset.
We design an automatic and scalable pipeline to collect high-quality AoT reasoning processes with the involvement of LLMs.
In addition to the AoT reasoning processes represented in natural language, the \ourdataset also includes AoT with programming language.
Such a hybrid training strategy could not only unleash the potential of the code use, but also allow the flexibility of preferring different reasoning processes for different reasoning problems~\cite{DBLP:journals/corr/abs-2309-05653}.
We use \ourdataset to finetune a wide range of language models.

We conduct exhaustive experiments on 23 unseen tasks from Big-Bench Hard~\cite{DBLP:conf/acl/SuzgunSSGTCCLCZ23}, a subset of the most challenging reasoning tasks of Big-Bench~\cite{DBLP:journals/corr/abs-2206-04615} that necessitates various reasoning abilities.
Experimental results show that AoT makes language models better reasoners.
Remarkably, models with AoT-finetuning achieve substantial improvements in both zero-shot and few-shot performance across various reasoning tasks, compared to those with CoT-finetuning.
Our findings highlight the potential of AoT in eliciting and training more effective models capable of reasoning with abstraction.

\section{Related Work}

\subsection{Chain-of-Thought Prompting}
The enlargement of the language model scale brings about emergent abilities including in-context learning~\cite{DBLP:journals/tmlr/WeiTBRZBYBZMCHVLDF22}.
To utilize LLMs for reasoning tasks, \citet{DBLP:conf/nips/Wei0SBIXCLZ22} propose CoT prompting, which extends in-context learning with step-by-step reasoning processes to elicit reasoning in LLMs. 
Subsequently, a substantial number of works based on CoT are proposed to further enhance LLMs' reasoning performance~\cite{DBLP:journals/corr/abs-2309-15402,DBLP:journals/corr/abs-2401-14295, DBLP:conf/nips/KojimaGRMI22,DBLP:conf/nips/YaoYZS00N23}. 
For instance, \citet{DBLP:journals/corr/abs-2211-12588} and \citet{DBLP:conf/icml/GaoMZ00YCN23} explore using codes to express the reasoning processes.
Most of these existing approaches focus on the content of the prompt (e.g., question decomposition~\cite{DBLP:conf/iclr/ZhouSHWS0SCBLC23}) and the external usage of CoT prompt (e.g., tree searching~\cite{DBLP:conf/nips/YaoYZS00N23}).
We focus on the format and internal structure of CoT prompts and propose the Abstraction-of-Thought format, which is complementary to existing approaches.
Our approach can potentially be combined with existing methods for further improvement.

\subsection{Training Language Models for Reasoning}
Although LLMs equipped with CoT prompts can achieve advanced reasoning capabilities, there is still a significant gap between open-source smaller-scale models and large models. 
To bridge this gap, a promising and popular approach is finetuning language models to learn reasoning~\cite{DBLP:journals/corr/abs-2402-13116}. 
A series of studies have found that finetuning models on data containing CoT reasoning processes could enhance the reasoning ability~\cite{DBLP:journals/corr/abs-2402-13116,DBLP:conf/nips/ZelikmanWMG22,DBLP:conf/acl/ShridharSS23,DBLP:conf/acl/HoSY23,DBLP:conf/icml/FuPOSK23,DBLP:journals/corr/abs-2307-02053,DBLP:conf/acl/HsiehLYNFRKLP23,DBLP:conf/acl/MagisterMAMS23,DBLP:journals/corr/abs-2210-06726,DBLP:journals/corr/abs-2305-13888,DBLP:journals/corr/abs-2309-05653}. 
The predominant way is to train models on instruction tuning datasets and their enhanced versions. 
For instance, the instruction tuning dataset FLAN collection~\cite{DBLP:conf/icml/LongpreHVWCTZLZ23} includes CoT data on a small subset of tasks to improve the model's performance under CoT prompts. 
CoT Collection~\cite{DBLP:conf/emnlp/KimJKJYSS23} further supplements the remaining tasks in FLAN with CoT reasoning processes, covering 1.84 million instances. 
Orca~\cite{DBLP:journals/corr/abs-2311-11045,DBLP:journals/corr/abs-2306-02707} enhances FLAN by prompting LLM with task-specific prompts, thereby training smaller models for cautious reasoning.
In this paper, we follow previous work to collect data based on FLAN for fair comparison.
We design the methodology for gathering AoT reasoning process and collect \ourdataset to facilitate better training of models for reasoning.

\subsection{Reasoning with Abstraction}
Previous works study different aspects of reasoning with abstraction, including entity abstraction~\cite{DBLP:conf/eacl/DurmeMS09,DBLP:conf/ijcai/SongWWLC11,DBLP:conf/aaai/GongZZ16}, event abstraction~\cite{DBLP:journals/corr/abs-2206-01532,DBLP:journals/corr/abs-2311-09174}, spatial-temporal abstraction~\cite{DBLP:conf/cvpr/0017JZZ21}, and conceptualization abstraction~\cite{DBLP:journals/corr/abs-2404-00205}.
\citet{DBLP:journals/corr/abs-2401-17464} propose to use abstract placeholders in the reasoning chain and call domain tools to supplement specific knowledge, thus allowing the model to use the tools effectively.
\revise{\citet{DBLP:journals/corr/abs-2306-17820} propose to transform questions into symbolic meta forms, but require complex semantic resolution and rely on specific entities.}
We explore how to improve the basic abstract reasoning of LMs from the perspective of reasoning formats and training data.

\begin{figure*}[t]
    \centering
    \includegraphics[width=.98\textwidth]{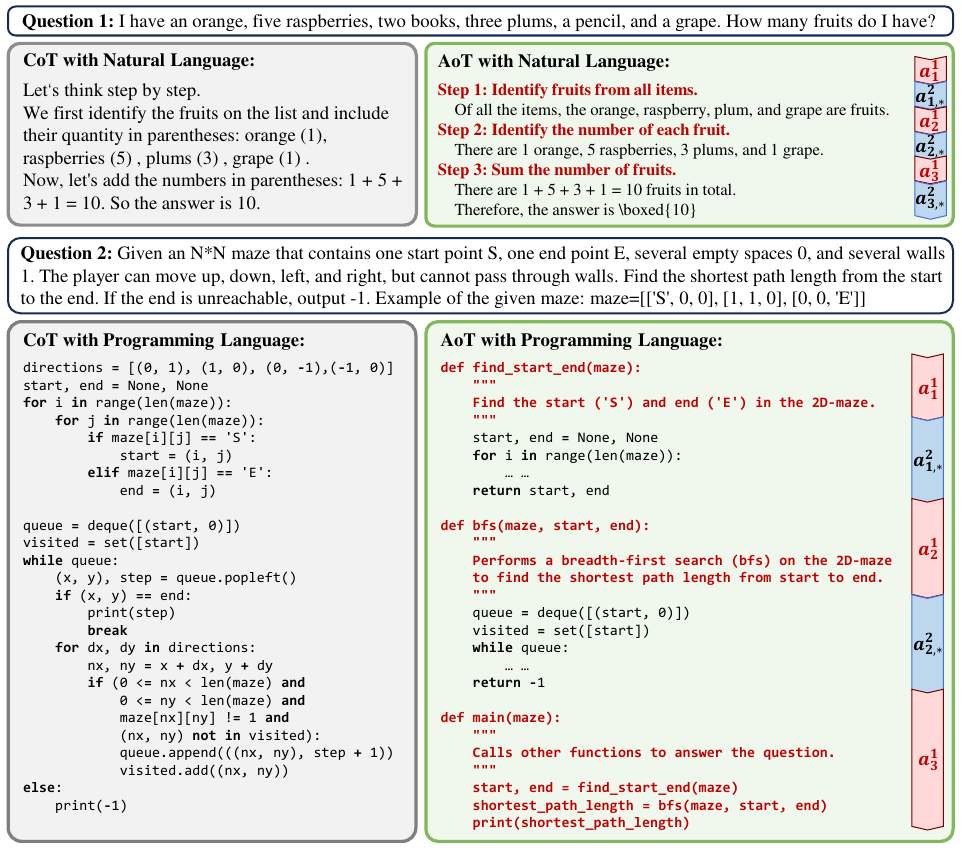}
    \caption{Illustration of Abstraction-of-Thought (AoT) format with natural language (upper half) and programming language (lower half).
    Unlike the unconstrained CoT, AoT explicitly requires that different levels of abstraction be presented in the reasoning process.
    Here are examples of two-level abstraction AoT.
    In AoT, the high-level parts (represented in {\DarkRed{\textbf{bold red}}}, i.e., \DarkRed{$\bm{a_{*}^{1}}$}) plan and organize the entire reasoning process from an abstract perspective, while low-level parts (i.e., $a_{*,*}^{2}$) carry out concrete and detailed reasoning steps. 
    The high-level parts are abstractions of the low-level parts, clarifying their functionality and objectives.
    For clarity, we omit some code snippets in AoT with programming language.}
    \label{fig:aot}
\vspace{-6mm}
\end{figure*}

\section{Abstraction-of-Thought (AoT)}

The Abstraction-of-Thought (AoT) reasoning format draws inspiration from the human application of abstract thinking to solve complex reasoning and planning problems~\cite{saitta2013abstraction,yang2012intelligent}.
In trying to solve a complex problem, a good strategy adopted by human reasoners is to proceed top-down.
They start from a rough abstract solution and then gradually refine it until a concrete solution is found. 
For instance, when faced with a complex programming task, seasoned programmers often commence by sketching out a rough algorithmic framework and identifying abstract function features. Subsequently, they progressively delve into the specifics of the code implementation.
Similarly, in argumentative discourse~\cite{DBLP:journals/coling/Cohen87}, the arguer often first presents the main claims, followed by detailed explanations and supportive evidence for these claims. 
Therefore, complex problem solutions frequently contain information at diverse abstraction levels, which serves to organize the entire reasoning process hierarchically. 
These high levels (e.g., functions in a coding solution or main claims in an argument) encapsulate the functionality and objectives of lower levels, thereby structuring the internal reasoning process.

We suggest that the explicit inclusion of such multiple levels of abstraction in the reasoning process could be the key to eliciting abstract reasoning in LMs and propose the AoT format (Figure~\ref{fig:aot}). 
\revise{Specifically, an $n$-step AoT reasoning process can be expressed as $\tau_{AoT} = a_{1}^{1} \circ a_{2}^{1} \circ \dots \circ a_{n}^{1}$. 
where $\circ$ denotes concatenation and $a_{i}^{j}$ is the $i$-th step in the $j$-th abstraction level.}
\revise{Each $a_{i}^{j}$ can be expanded to $a_{i,1}^{j+1} \circ \dots \circ a_{i,m_{ij}}^{j+1}$, which includes $m_{ij}$ steps belonging to the $(j+1)$-th abstraction level.}
The high-level part (e.g., $a_i^{1}$) provides the abstract solution (e.g.,  deriving the general quadratic formula for solving a specific quadratic equation in Figure~\ref{fig:intro}).
It focuses on the essence of the problem and ignores inessential details.
The low-level part (e.g., $a_i^{j \ge 2}$) further refines it until concrete operations and reasoning can be performed.
It contains more details and specific conditions.
\revise{AoT can be regarded as a constrained variant of CoT, constraining the internal structure of the reasoning process from the perspective of abstract reasoning.}

\revise{As a preliminary exploration into whether abstract content aids in the reasoning of LLMs, this paper focuses on the two-level AoT.
Compared to the unconstrained CoT, the two-level AoT already includes an additional level of abstraction, which is sufficient for our investigation.}
\revise{We define the specific AoT format on both natural and programming language, as shown in Figure~\ref{fig:aot}.
For natural language, AoT requires (1) a clear division of the reasoning process into steps, with ``Step i'' as the beginning of the $i$-th step; (2) stating the abstract purpose of the step (as the first level of abstraction $a_{*}^{1}$) followed by the specifics in each reasoning step (as the second level of abstraction $a_{*,*}^{2}$); and (3) placing the reasoning result in ``\texttt{\textbackslash boxed\{\}}''.
For programming language, AoT requires (1) explicitly dividing the code into several pieces, each of which is a function or class; (2) annotating the functions or classes to explain their functionality (as $a_{*}^{1}$); and (3) calling other functions in the main function to solve the problem.
The code within the main function is also considered the first level, as it forms the abstract plan for solving the problem.}
\revise{Note that there may be multiple steps of $a_{i,*}^{2}$ under $a_{i}^{1}$, depending on the difficulty of the problem.
For $a_{i,*}^{2}$, we consider a line as a step, i.e., different $a_{i,*}^{2}$ are divided by line breaks.}

\section{The \ourdataset}

\subsection{Overview}

To align LMs to the AoT reasoning format, we collect the \ourdataset for finetuning LMs.
\ourdataset comprises reasoning processes in AoT format, characterized by the following features:
(1) \textbf{Hybrid reasoning strategy.} 
\ourdataset incorporates reasoning processes expressed in both natural language and programming language. 
Inspired by previous work~\cite{DBLP:conf/icml/GaoMZ00YCN23,DBLP:journals/corr/abs-2211-12588}, solutions to some complex problems are more appropriately to be expressed with programming language.
This approach not only facilitates the use of external modules as tools to improve reasoning accuracy~\cite{DBLP:journals/corr/abs-2401-00812}, but also enhances the faithfulness of the reasoning process~\cite{DBLP:conf/ijcnlp/LyuHSZRWAC23}. 
Models trained on such hybrid reasoning strategies can flexibly choose the suitable reasoning approach based on the type of test question, thereby handling more complex reasoning problems.
(2) \textbf{Broad task coverage.}
\ourdataset is built upon an instruction-tuning dataset that covers a wide range of tasks. 
This allows the \ourdataset to encompass questions from various tasks, rather than being confined to a specific domain or task.
(3) \textbf{Scalability.} 
\ourdataset is collected through an automated pipeline, which allows it to scale up without relying on the efforts of human annotation.

\subsection{Source Dataset Selection}
We follow previous work~\cite{DBLP:conf/emnlp/KimJKJYSS23,DBLP:journals/corr/abs-2311-11045} to use FLAN Collection~\cite{DBLP:conf/icml/LongpreHVWCTZLZ23} as our source dataset.
FLAN Collection includes diverse NLP datasets sourced from P3~\cite{DBLP:conf/iclr/SanhWRBSACSRDBX22}, Super-NaturalInstructions~\cite{DBLP:conf/emnlp/WangMAKMNADASPK22}, Flan~\cite{DBLP:conf/iclr/WeiBZGYLDDL22}, and additional dialogue and code datasets.
We follow~\citet{DBLP:conf/emnlp/KimJKJYSS23} to exclude datasets whose data are not publicly accessible and datasets with an excessive number of tokens.
Ultimately, we focus on 216 datasets that are consistent with the CoT Collection~\cite{DBLP:conf/emnlp/KimJKJYSS23}.
\revise{These tasks cover different types of tasks in many domains (Appendix~\ref{sec:appendix_data}).}
We manually divide the 216 datasets into two parts, \textbf{AoT-Text} (203 datasets that are more suitable to be solved in natural language) and \textbf{AoT-Code} (13 datasets that are more suitable to be solved in programming language).
We utilize a proportional stratified sampling method to sample 400k instances from the original data for the subsequent AoT response generation.
Details of the dataset division can be found in Appendix~\ref{sec:appendix_data}.

\subsection{AoT Response Generation}
\revise{While manually annotating the AoT reasoning process for all instances could yield higher quality, it is time-consuming and labor-intensive, and thus difficult to scale up.}
We adopt a method of synthesizing data with LLMs~\cite{DBLP:conf/emnlp/KimJKJYSS23,liu2024best} to automate the process of collecting AoT responses. 
We first manually create instructions and 3 demonstrations, to exploit the instruction-following and in-context learning capabilities of LLM for generating AoT responses (details can be found in Appendix~\ref{sec:appendix_data}). 
To minimize the difficulty of generation, we only consider 2 levels of abstraction\footnote{\revise{Our data collection process might not guarantee to find the optimal abstraction solution for each question, since the optimal abstraction level may differ across various questions, and a single question may have multiple abstract solutions. We focus on investigating whether valid abstraction can help language models reasoning while deprioritizing the collection of optimal abstraction solutions.}}.
We designed two types of prompts for the datasets in {AoT-Text} and {AoT-Code}, respectively.
The correct answer to the question is included in the prompt to help the model focus on the generation of the reasoning process.
We use GPT-3.5-Turbo as our back-end LLM and generate with greedy decoding.
Since we are more concerned about the impact of the reasoning format on the model's reasoning ability, we do not meticulously design different demonstrations for each dataset like previous work~\cite{DBLP:conf/emnlp/KimJKJYSS23, DBLP:journals/corr/abs-2311-11045}. 
We use Python as the programming language.

\begin{table}[t]

\small
\centering
\resizebox{\columnwidth}{!}{%
\begin{tabular}{@{}lccc@{}}
\toprule
           & \textbf{AoT-Text} & \textbf{AoT-Code} & \textbf{Total} \\ \midrule
Number of Samples &  173,100  &  175,463   &  348,563     \\
Avg. Question Length  &   179.6  &   98.9  &   139.0    \\ 
Avg. AoT Response Length &  144.6  &  172.2  &  158.5    \\ 
\bottomrule
\end{tabular}%
}
\caption{Statistics of \ourdataset}
\label{tab:statistic}
\vspace{-5mm}
\end{table}

\subsection{Data Validation and Filtering}

After generating the AoT response, we perform validation and filtering to ensure high quality.
For AoT in natural language, we examine whether the answers predicted in the response are consistent with gold answers.
To prevent degeneration where different inference steps describe the same content, we stipulate that there should not be excessive similarity between different steps.
Specifically, we calculate the Jaccard similarity of words between different steps and require it to be below a threshold of 0.5.
For AoT in the programming language format, we execute the code provided in the response and check whether it correctly prints or returns the gold answer.
For instances that fail to meet the requirements, we ask the LLM to regenerate 10 times (with a temperature parameter of 0.7).
We retain the first response that meets the requirement. If none of the 10 responses meet the requirement, we discard the instance.
After the filtering process, we retain 348k instances.
We randomly sampled 100 examples (50 in natural language and 50 in programming languages) and manually checked the quality of the AoT responses.
We find that 96\% of AoT responses are valid.
Table~\ref{tab:statistic} reports the statistics and Appendix~\ref{sec:appendix_data} lists some samples of \ourdataset.

\section{Experiments}

\noindent $\bullet$ \textbf{Evaluation Dataset.}
We evaluate with the challenging reasoning benchmark BIG-Bench Hard (BBH)~\cite{DBLP:conf/acl/SuzgunSSGTCCLCZ23}, which is the most challenging subset of BIG-Bench~\cite{DBLP:journals/corr/abs-2206-04615}.
BBH consists of 23 tasks that are specifically selected for their difficulty for LMs.
BBH covers a wide range of reasoning challenges, including semantic reasoning (e.g., Movie Recommendation), numerical reasoning (e.g., Multi-Step Arithmetic), logical reasoning (e.g., Logical Deduction), 
and combinations of some of these abilities (e.g., Object Counting).
Furthermore, the FLAN Collection takes BBH as \textit{Held-Out} tasks, which ensures that our finetuning process does not access the evaluation data.
We report the average accuracy across 12 NLP tasks (\textbf{NLP}), 11 algorithm tasks (\textbf{Alg}), and all 23 tasks (\textbf{All}), respectively.
Details about BBH are in Appendix~\ref{sec:appendix_bbh}.

\noindent $\bullet$ \textbf{Setting and Baselines.}
We finetune LMs with \ourdataset and evaluate their reasoning ability.
Following previous work~\cite{DBLP:conf/emnlp/KimJKJYSS23,DBLP:journals/corr/abs-2311-11045}, we focus our evaluation primarily on the zero-shot setting.
The zero-shot setting represents the realistic scenario, as in practical applications we do not have prior knowledge (e.g., few-shot demonstrations) about the test questions.
We compare the CoT-finetuned version (e.g., Llama-3-8B-\textbf{CoT}) and AoT-finetuned version (e.g., Llama-3-8B-\textbf{AoT}) of LMs.
For CoT-finetuning, we replace the reasoning processes in \ourdataset with the CoT rationales provided by the CoT Collection~\cite{DBLP:conf/emnlp/KimJKJYSS23}.
We also report the performance of the instruction-finetuned LMs (e.g., Llama-3-8B-Instruct) as a reference.

\noindent $\bullet$ \textbf{Models.}
We consider a range of common open source pre-trained language models, including Llama-2~\cite{DBLP:journals/corr/abs-2307-09288}, CodeLlama~\cite{DBLP:journals/corr/abs-2308-12950}, Llama-3~\cite{llama3modelcard}, Mistral~\cite{DBLP:journals/corr/abs-2310-06825}.
We also report the performance of GPT-3.5-Turbo-0125~\cite{GPT-3.5} and GPT-4-0613~\cite{DBLP:journals/corr/abs-2303-08774}.
Details about the models can be found in Appendix~\ref{sec:appendix_model}.

\noindent $\bullet$ \textbf{Implementation Details.}
We train all models with Megatron-LM~\cite{DBLP:journals/corr/abs-1909-08053}.
We use a learning rate of 2e-6 for 1 epoch by default. 
We set the global batch size to 128 and use a cosine decay scheduler.
We use greedy decoding for all results, with the maximum sequence length set to 2,048.
For the response in natural language, we extract the contents within ``\texttt{\textbackslash boxed\{\}}'' as the predicted answer. 
For the response in the programming language format, we execute the program and take the printed output as the predicted answer.

\subsection{Zero-Shot Performance}

\input{tables/table_zero_shot}

Table~\ref{tab:Zero-shot} presents the zero-shot BBH performance of LMs finetuned in different ways.
The AoT-finetuned models demonstrate a remarkable performance enhancement compared to the baselines.
For example, in comparison to Llama-3-8B-CoT, Llama-3-8B-AoT achieves an absolute improvement of +9.7\%.
This trend is consistent across all models studied, which suggests that aligning models with the AoT reasoning format could enhance their reasoning ability more effectively than aligning them to the CoT.
AoT guides models to think from a more abstract perspective first, addressing the essence of the problem at a higher level, which potentially enhances their reasoning ability.

It is worth noting that the AoT-finetuned models displayed more substantial improvements in algorithmic tasks than in NLP tasks.
The algorithmic tasks, which require capturing the internal reasoning rules of the questions without relying on external knowledge, pose a greater challenge to the model's reasoning ability~\cite{DBLP:conf/acl/SuzgunSSGTCCLCZ23}.
As a result, the performance on algorithmic tasks is usually lower.
In contrast, NLP tasks depend not only on reasoning ability but also on external knowledge, where LMs could face the bottleneck in external knowledge.
AoT-finetuning guides the model to carry out reasoning with abstraction, with a focus on strengthening the reasoning ability, thus yielding a more noticeable improvement in algorithmic tasks.
We further discuss the performance of subtasks (\ref{sec:appendix_by_tasks}), case study (\ref{sec:appendix_case_study}), and computational cost comparison (\ref{sec:appendix_computational_cost}) in the Appendix.

\input{tables/table_few_shot}

\subsection{Few-Shot Performance}

We also evaluate the effect of AoT under the few-shot setting, the standard setting proposed by~\citet{DBLP:conf/acl/SuzgunSSGTCCLCZ23}.
For few-shot CoT demonstrations, we use three questions and their CoT rationales which are provided by the official repository.
For the AoT prompt, we employ the same questions and manually create the AoT rationales. 
Consistent with our \ourdataset, we use the Python program to express the reasoning process for some tasks.
Prompts can be found in Appendix~\ref{sec:appendix_few_shot}.

Table~\ref{tab:Few-shot} shows the results. 
For models that have not been finetuned, using prompts in AoT format achieves remarkable performance improvement compared to those in the CoT format.
For example, the pre-trained Llama-3-8B achieves an absolute improvement of 15.1\% with the AoT prompting.
This suggests that AoT could be more effective in stimulating the reasoning ability of pre-trained LMs.
\revise{Furthermore, by aligning the pre-trained models to AoT with the \ourdataset, the models demonstrate improved performance under AoT prompts, validating the effectiveness of our \ourdataset.}
Further discussions including the few-shot performance of instruction-finetuned models are in Appendix~\ref{sec:appendix_few_details}.

\input{tables/table_aot2cot}

\subsection{Ablation Study on Reasoning Format}
In preceding experiments, we utilize the CoT Collection~\cite{DBLP:conf/emnlp/KimJKJYSS23} for the CoT-finetuning. 
CoT Collection differs from our \ourdataset in two confounding factors, besides the reasoning format: 
(1) A different LLM is used during data collection; 
(2) \ourdataset additionally employs the hybrid reasoning strategy, representing reasoning both in text and code. 
To verify the role of the AoT format, we conduct an ablation study to attempt to eliminate the influence of these confounding factors. 
\revise{We construct a new training dataset, \aottocot, which uses the same LLM (i.e., GPT-3.5-Turbo) as the \ourdataset to collect data and also adopts the hybrid reasoning strategy. }
Specifically, we prompt the LLM to convert the reasoning processes of \ourdataset from AoT into CoT while keeping the main reasoning content the same.
\revise{Meanwhile, we also compare with \ourcot, where we use the same back-end LLM to generate CoT directly, adopting the same hybrid reasoning strategy and filtering approach as for AoT data.}
Considering the computational cost of the LLM, we sample 10k data from \ourdataset, and carry out the ablation experiment on these data. 
\revise{We finetune the models on these same 10k questions, but with reasoning processes in different formats:
(1) \textbf{CoT} from CoT Collection,
(2) \textbf{\ourcot},
(3) \textbf{AoT2CoT},
and (4) \textbf{AoT} from \ourdataset.}
More implementation details can be found in the Appendix~\ref{sec:appendix_aot2cot}.

As shown in Table~\ref{tab:aot2cot},
\revise{among all formats of training data, AoT achieves the best results over the CoT format.
With the same back-end LLM and hybrid reasoning strategy, AoT still outperforms AoT2CoT and \ourcot.
This demonstrates that the main factor contributing to our method's improvement is the reasoning format. 
}

\subsection{Ablation Study on Training Data Scale}
\begin{figure}
    \centering
    \includegraphics[width=0.9\columnwidth]{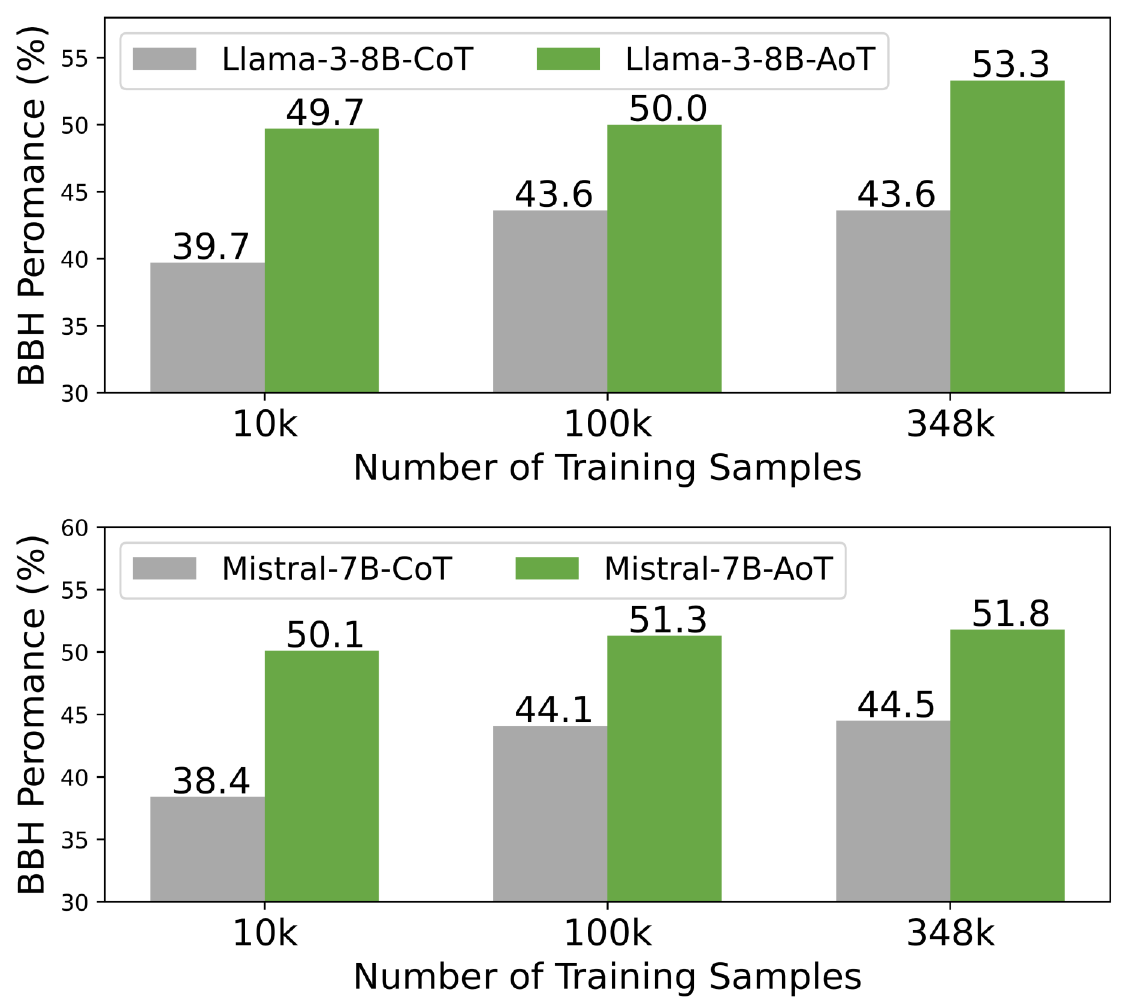}
    \caption{Zero-shot BBH performance of models trained with different numbers of training samples.}
    \label{fig:ablation_scale}
\vspace{-3mm}
\end{figure}

To investigate the impact of the training data scale, we train models with different numbers of training samples.
As shown in Figure~\ref{fig:ablation_scale}, the AoT-finetuned models achieve a steady improvement as the scale of training data increases.
Across all scales, AoT-finetuned models outperform CoT-finetuned models.
Moreover, finetuning using 10k AoT data can yield desirable performance, even better than the case of finetuning using 348k CoT data.
This demonstrates both the effectiveness and efficiency of our \ourdataset.

\subsection{Response Analysis}
We analyze the responses of AoT-finetuned models from the following perspectives.

\noindent $\bullet$ \textbf{Usage Rate (UR)}: The proportion of the text/code being utilized among all responses.

\noindent $\bullet$ \textbf{Format Correctness (FC)}: The proportion of responses that contain/output predicted answers and follow the AoT format among text/code responses.

\noindent $\bullet$ \textbf{Answer Correctness (AC)}: The proportion of responses whose predicted answers are correct among text/code responses.

Table~\ref{tab:ablation_model_analysis} presents the results of the responses on BBH in the zero-shot setting.
\revise{We can find that models prefer to reason with natural language for most problems, rather than programming language. 
For example, only 26.6\% of responses from Llama-3-8B-AoT use code.
Moreover, reasoning with code often achieves higher accuracy than reasoning with text. 
The accuracy of code responses from CodeLlama-7B-AoT reaches 61.1\%, whereas text responses only have an accuracy of 43.6\%. }

\input{tables/table_model_analysis}

\subsection{Error Analysis}

To better understand the failure modes and future challenges, we randomly sample 100 problems on which Llama-3-8B-AoT fails and manually annotate their error types.
Mostly, incorrect reasoning steps constitute the main cause of errors (38\%). 
The model also suffers from lack of necessary knowledge (16\%), misunderstanding of tasks (15\%), and hallucination (12\%).
The reasoning process sometimes trivially repeats the conditions (12\%) or fails to be executed (5\%).
Finally, a smaller percentage of correctly predicted answers are misjudged by automated indicators (5\%).
We present detailed error definitions, examples, and proportions in Appendix~\ref{sec:appendix_case_study} Table~\ref{tab:table_error}.

\section{Conclusion}

This paper explores how to elicit language models to perform reasoning with abstraction, from the perspective of the reasoning format.
We propose a novel structured reasoning format, AoT, which explicitly requires multiple levels of abstraction to be included in the reasoning process.
We construct the \ourdataset, containing 348k high-quality AoT reasoning processes, for fine-tuning models to align them to the AoT format.
Experimental results show that AoT-finetuned language models can achieve advanced reasoning performance compared to baseline approaches.

\section*{Limitations}

In this paper, we explore how to elicit language models to perform abstract reasoning from the perspectives of the reasoning format, thereby improving their performance on reasoning tasks. 
While considerable progress has been made, there are still room for improvement and future research areas worth exploring.

First, due to resource constraints, this paper has not yet explored how to enhance the model's abstract reasoning ability during the pre-training stage. 
In our experiments, it can be observed that there is a substantial difference in performance among different pre-trained models after AoT-finetuning. 
This is due to the issue that the finetuning stage might be difficult to fundamentally enhance the intrinsic capabilities of the model~\cite{DBLP:journals/corr/abs-2204-05862,DBLP:conf/nips/Ouyang0JAWMZASR22}. 
Therefore, the bottleneck to fundamentally endow models with the ability for abstract reasoning might lie in the pre-training stage. 
How to inject the ability for abstract reasoning into language models during the pre-training stage presents an intriguing direction for future research. 

Second, limited by the cost of API calls, we do not conduct more analysis on the back-end LLM used for data collection. 
The impact of using a more powerful LLM for data collection or collecting data on a larger scale for AoT-finetuning deserves further exploration.

\revise{Third, this paper focuses the evaluation of the model's reasoning ability on the BBH benchmark, which contains 23 test tasks that require various reasoning abilities.
However, each task of BBH has its own problem-solving pattern and may favor logical and symbolic reasoning over other benchmarks. 
Further exploration of generalizability on more realistic and complex reasoning problems is a future research direction.}

This paper follows the ACL Code of Ethics. To the best of our knowledge, our work is foundational research, and we do not find obvious risks related to fairness considerations, privacy considerations, malicious harmful effects, or environmental impact.

\section*{Acknowledgements}
\revise{We appreciate the anonymous reviewers for their insightful comments.}
\revise{This work is supported by National Science and Technology Major Project (No. 2022ZD0114903), the Natural Science Fundation of China (NSFC. No. 62476149) and the Guoqiang Institute of Tsinghua University (No. 2020GQG0005).}

\bibliography{anthology,custom}

\newpage
\clearpage
\appendix

\section{Implementation Details}
\label{sec:appendix_details}

\subsection{\ourdataset}
\label{sec:appendix_data}

We follow CoT Collection~\cite{DBLP:conf/emnlp/KimJKJYSS23}\footnote{\href{https://huggingface.co/datasets/kaist-ai/CoT-Collection}{https://huggingface.co/datasets/kaist-ai/CoT-Collection}} to preprocess the FLAN Collection~\cite{DBLP:conf/icml/LongpreHVWCTZLZ23} data as our source data.
The CoT Collection contains 1.84M samples from 216 tasks.
\revise{These tasks cover different types of tasks in different domains. 
For example, dataset ``commonsenseqa'' in the collection is a ``Question Answering'' task in the commonsense domain, while dataset ``gsm8k'' is a ``Grade School Math Word Problems'' in the math domain.
Our \ourdataset is built on these diverse datasets and already covers different tasks in different domains as well, rather than being designed only for a specific domain.
}

\revise{Table~\ref{tab:aot_task_detail} at the end of the Appendix gives all the dataset names and their corresponding tasks.
We follow the FLAN Collection to determine the tasks corresponding to the datasets.
Considering the large number of datasets, we have opted not to display examples for each one in this document. 
Please refer to the FLAN Collection (\url{https://github.com/google-research/FLAN/tree/main/flan/v2}) if you are interested in the specific dataset examples.
To demonstrate the diversity of tasks, we list the task categories and their distribution below. The number in parentheses indicates the dataset for which this task is relevant.}

\input{tables/table_aot_task_counter}

Each sample of CoT Collection consists of a question, an answer, and a CoT rationale.
We discard the CoT rationale and use the GPT-3.5-Turbo to generate the AoT reasoning process based on the question and the correct answer.
We manually divided the 216 datasets into two parts, \textbf{AoT-Text} (203 datasets that are more suitable to be solved in natural language) and \textbf{AoT-Code} (13 datasets that are more suitable to be solved in programming language).
The AoT-Text consists of the following tasks:

\vspace{+2mm}
\noindent
\begin{minipage}{\columnwidth}
\small
adversarial\_qa@droberta, olid, ai2\_arithmetic\_questions, natural\_questions, question\_\&\_answer\_zre, qanta, health\_fact, quail, jeopardy, jigsaw, detoxifying\_lms, poki, qnli, cb, google\_wellformed\_query, subjqa, multi\_woz\_v2, casino, task\_master\_input\_inversion, babi, wiki\_movies, ddo, anli\_r1, craigslist\_bargains, ohsumed, strategyqa, adversarial\_qa@dbidaf, cad, semeval\_2018\_task3, emo, overruling, hippocorpus, qed, diplomacy\_detection, piqa, smcalflow, super\_glue@record, schema\_guided\_dstc8, ai2\_arc@ARC-Easy, nlu\_asdiv\_dataset, ruletaker, news\_headlines\_dataset\_for\_sacrasm\_detection, com2sense, civil\_comments, circa, quartz, diqa, semeval\_2019\_task\_10, hybridqa, evaluation, ecqa, storycommonsense, miscellaneous, snli, clariq, blimp, financial\_phrasebank, hatexplain, hope\_edi, numersense, x\_csr, xcsr, qa\_srl, mcscript, mwsc, persent, trivia\_qa, hate\_speech\_offensive, coached\_conv\_pref, scitail, drop, rte, anli\_r3, qrecc, ms\_marco, quac, wikitext\_103, nlg\_bias, mutual, gwsd, yahoo\_answers\_topics, essential, swag, torque, wiki\_dialog\_input\_inversion, cola, winowhy, disfl\_qa, roc\_stories, semeval\_2020\_task\_7, codah, mocah, atomic, crows\_pair, mnli, tweetqa, scruples, conv\_ai\_2, stereoset, break, duorc@SelfRC, dialogre, ambigqa, iirc, miam, pubmed\_qa, deal\_or\_no\_dialogue, ai2\_arc@ARC-Challenge, coda\_19, spolin, wiki\_hop, hateeval, timetravel, duorc@ParaphraseRC, recepie\_nlg, kilt\_tasks@hotpotqa, curated\_from\_stack\_overflow\_\_\_english, sciq, freebase\_qa, squad\_v2, help!\_need\_advice\_on\_identifying\_advice, bless, squad\_v1, task\_master, sbic, quoref, com\_qa, wnli, haspart\_kb, personachat, argkp, ethos, open\_pi, race@high, proto\_qa, sarcasm\_in\_twitter, web\_questions, abductive\_nli, curiosity\_dialogs, imppres, race@middle, adversarial\_qa@dbert, eurlex, head\_qa, defeasible\_nli\_atomic, equity\_evaluation\_corpus, qrecc\_input\_inversion, wiki\_dialog, eqasc, bard, wiqa, dream, liar, anli\_r2, scitailv1.1, tellmewhy, cod3s, dstc, indian\_food\_101, aquamuse, glucose, social\_i\_qa, air\_dialogue, missing, narrativeqa, scitldr, mrqa, meta\_woz, go\_emotions, casehold, scifact, super\_glue@boolq, ade\_corpus\_v2, dailydialog, starcon, commonsenseqa, openbookqa, quarel, propara, event2mind, inquistive, tom\_qa, wiki\_qa, cosmos\_qa, afs, medical\_question\_pair\_dataset, creak, yoruba\_bbc\_topics, semeval\_2020\_task4, xl\_wic, super\_glue@multirc, opp\_115, esnli, grailqa, root09, qasper, ropes, gooaq, cos\_e, perspectrum, xquad, trianglecopa, mctaco
\end{minipage}
\vspace{+2mm}

The AoT-Code consists of the following tasks:

\vspace{+2mm}
\noindent
\begin{minipage}{\columnwidth}
\small
aqua, big\_bench, cfq\_mcd1, conala, gsm8k, leetcode, math\_dataset, math\_qa, mathmatics\_dataset, mathqa, prost, svamp, synthetic
\end{minipage}
\vspace{+2mm}

We utilize a proportional stratified sampling method to sample 200k instances from the AoT-Code datasets and 200k instances from the AoT-Text datasets.
The proportional stratified sampling method ensures that the proportion of each dataset is the same before and after sampling, thus maintaining the overall data distribution.

Listing~\ref{lst:prompt_aot_collect} shows the system prompts we use for collecting AoT responses.
We use different prompts for AoT-Text and AoT-Code and clearly state the requirements of AoT format.
We also include 3 demonstrations in the prompt.

\ourdataset retains 348k high quality data after filtering.
The average number of steps for responses in AoT-Text and AoT-Code is 2.8 and 2.2, respectively.
We randomly sample 3 samples from each of AoT-Text and AoT-Code and present them in Listing~\ref{lst:Aot_Collection_Samples}.
\ourdataset is in English.

\begin{lstlisting}[caption={Prompts for collecting AoT responses. We show the system prompt and one of the three demonstrations.}, label={lst:prompt_aot_collect}]
Prompt for Collecting AoT Responses with Natural Language

** System Prompt **
You are a powerful reasoning model. I will present you with a question and the correct answer. Your task is to output the intermediate reasoning process to meet the following requirements. 
- The reasoning process contains several reasoning steps that can support the correct answer. 
- Each reasoning step should begin with a abstract high-level plan, followed by a detailed step-by-step implementation. 
- The high-level plans should be tailored to the specific question. They may encompass, but are not limited to: extracting key information from the question, sub-questions decomposed from the target question, analyzing certain aspects of the question, evaluating different options and possibilities, recalling relevant knowledge and rules, applying knowledge and rules for reasoning, comparing different content, concluding and answering. 
- Each step should aim to draw new conclusions and not merely repeat the question or the content of the previous step. 
- Conclude the reasoning process by stating the final answer as "The answer is \boxed{answer}". 

** Demonstration Input **
Which sentence has the correct adjective order: Options:
(A) Indonesian triangular obnoxious gold red shoe
(B) obnoxious triangular red Indonesian gold shoe
Correct Answer: (B)

** Demonstration Output **
Step 1: Recall the order of adjectives in English.
The generally accepted order of adjectives in English is: 
quantity, quality, size, age, shape, color, origin, material, purpose.

Step 2: Analyze each option according to the adjective order rule.
(A) Indonesian (origin) triangular (shape) obnoxious (quality) gold (material) red (color) shoe (noun)
(B) obnoxious (quality) triangular (shape) red (color) Indonesian (origin) gold (material) shoe (noun)

Step 3: Compare the options.
Option (A) has the incorrect order because "origin" comes before "shape" and "quality", and "material" comes before "color". 
Option (B) follows the correct order: quality, shape, color, origin, material.

Therefore, the answer is \boxed{(B)}.

================================================
Prompt for Collecting AoT Responses with Programming Language

** System Prompt **
You are a powerful reasoning model and Python coder. I will present you with a question and the correct answer. Your task is to output a Python program to meet the following requirements. 
- The code can answer the question and print the correct answer. 
- Divide the code into several functional blocks. Each functional block implements a simple function. 
- Describe what the function does in the comment lines following the function name. Comments should be as concise as possible. 
- Define a `main` function to call other functions to answer the question. 
- The main function should contain arguments transformed by the conditions in the question. The other functions should be as generic as possible. 

** Demonstration Input **
In this task, answer 'Yes' if the frequency of the given word in the two sentences is equal, else answer 'No’. Sentence1: 'a little boy that is holding a skateboard', Sentence2: 'this photo displays a messy desk that has two laptops on it among many other things'. Is the frequency of the word 'a' in two sentences equal?
Correct Answer: No

** Demonstration Output **
```python
def count_word_frequency(sentence, word):
    """
    Count the frequency of a given word in a sentence.
    """
    words = sentence.split()
    count = words.count(word)
    return count

def compare_word_frequency(sentence1, sentence2, word):
    """
    Compare the frequency of a given word in two sentences.
    """
    count1 = count_word_frequency(sentence1, word)
    count2 = count_word_frequency(sentence2, word)
    return count1 == count2

def main():
    """
    The main function that calls the other functions and prints the answer.
    """
    sentence1 = 'a little boy that is holding a skateboard'
    sentence2 = 'this photo displays a messy desk that has two laptops on it among many other things'
    word = 'a'
    is_equal = compare_word_frequency(sentence1, sentence2, word)
    print('Yes' if is_equal else 'No')

main()
```
\end{lstlisting}

\begin{lstlisting}[caption={Samples of \ourdataset.}, label={lst:Aot_Collection_Samples}]
=================== Sample 1 =================== 
** Input **
What requires nutrients to grow?
Choices:
- birds
- metals
- computers
- dirt

** Output **
Step 1: Recall the definition of nutrients.
Nutrients are substances that are essential for the growth, development, and maintenance of living organisms.

Step 2: Analyze each option and determine which one requires nutrients to grow.
- Birds: Birds are living organisms and require nutrients to grow.
- Metals: Metals are inanimate objects and do not require nutrients to grow.
- Computers: Computers are machines and do not require nutrients to grow.
- Dirt: Dirt is a mixture of minerals, organic matter, and other substances, but it does not require nutrients to grow.

Step 3: Compare the options and identify the correct answer.
Among the given choices, only birds require nutrients to grow.

Therefore, the answer is \boxed{birds}.

=================== Sample 2 =================== 
** Input **
Two analogies that relate objects to the associated rooms is given in the form "A : B. C : ?". "A : B" relates object A to room B. Your task is to replace the question mark (?) with the appropriate room for the given object C, following the "A : B" relation.

bookshelf : library. car : ?

** Output **
Step 1: Identify the relationship between the object and the room in the given analogy.
In the analogy "bookshelf : library", the bookshelf is typically found in the library.

Step 2: Apply the same relationship to the second part of the analogy.
Since a car is typically found in a garage, we can replace the question mark with "garage".

Therefore, the answer is \boxed{garage}.

=================== Sample 3 =================== 
** Input **
I have a new situation: Big town and Fair town both used petroleum products. Big town used them to make plastics and related materials, but didn't use it for gas, because they were environmentally conscious and used only electric vehicles for transportation. Fair town used petroleum for fuel mostly.

But I can use this background: A major use for petroleum products is fuel for cars, trucks, airplanes, trains, and other vehicles. The chemical used are usually a mixture of compounds containing several carbon atoms in a chain. When the material is ignited, a massive amount of gas is created almost instantaneously. This increase in volume will move the pistons in an internal combustion engine to provide power. A jet plane works on a similar principle. Air comes into the front of the engine and mixes with the jet fuel. This mixture is ignited and the gases formed create a lot of pressure to push the plane forward. The idea can be seen even more clearly in the case of a rocket launch. The ignition of the fuel (either solid-state or liquid) creates gases produced under great pressure that pushes the rocket up.

What is an answer for this question: Which town used more fuel for internal combustion engines?

** Output **
Step 1: Extract key information from the situation.
Big town uses petroleum products to make plastics and related materials, but not for gas as they use electric vehicles for transportation. Fair town uses petroleum mostly for fuel.

Step 2: Recall the background information.
The background information states that a major use for petroleum products is fuel for vehicles with internal combustion engines.

Step 3: Apply the background information to the situation.
Since Big town uses electric vehicles, they do not use petroleum for fuel in internal combustion engines. On the other hand, Fair town uses petroleum mostly for fuel, which implies they use it for internal combustion engines.

Therefore, the answer is \boxed{Fair}.

=================== Sample 4 =================== 
** Input **
In this task, you need to answer the given multiple-choice question on geometry. Classify your answers into 'a', 'b', 'c', 'd', and 'e'.

Problem: the radius of the two circular fields is in the ratio 4 : 5 the area of the first field is what percent less than the area of the second ? 
Options: a ) 42 % , b ) 39 % , c ) 38 % , d ) 36 % , e ) 26 %

** Output **
```python
def calculate_area(radius):
    """
    Calculate the area of a circle given its radius.
    """
    area = 3.14159 * radius**2
    return area

def calculate_percent_difference(area1, area2):
    """
    Calculate the percentage difference between two areas.
    """
    difference = area2 - area1
    percent_difference = (difference / area2) * 100
    return percent_difference

def main():
    """
    The main function that calls the other functions and prints the answer.
    """
    radius1 = 4
    radius2 = 5
    area1 = calculate_area(radius1)
    area2 = calculate_area(radius2)
    percent_less = calculate_percent_difference(area1, area2)

    options = {
        'a': 42,
        'b': 39,
        'c': 38,
        'd': 36,
        'e': 26
    }

    for option, percent in options.items():
        if percent == percent_less:
            print(option)
            break

main()

```

=================== Sample 5 =================== 
** Input **
In this task, you need to replace a letter in the sentence with another given letter.

Sentence: 'a red plus sign over a toilet, near a sink'. Replace the letter 'r' with 'o' in the sentence.

** Output **
```python
def replace_letter(sentence, old_letter, new_letter):
    """
    Replace a letter in a sentence with another given letter.
    """
    replaced_sentence = sentence.replace(old_letter, new_letter)
    return replaced_sentence

def main():
    """
    The main function that calls the other functions and prints the answer.
    """
    sentence = 'a red plus sign over a toilet, near a sink'
    old_letter = 'r'
    new_letter = 'o'
    replaced_sentence = replace_letter(sentence, old_letter, new_letter)
    print(replaced_sentence)

main()

```

=================== Sample 6 =================== 
** Input **
In this task, you are given an input list. A list contains several comma-separated items written within brackets. You need to return the count of all the numerical elements in the given list. Return 0 if no numerical element in the list.

['D', '9987', '4565', 'j', 'V', 'S', '1033', '7319', 'S']

** Output **
```python
def count_numerical_elements(lst):
    """
    Count the number of numerical elements in a given list.
    """
    count = 0
    for item in lst:
        if item.isdigit():
            count += 1
    return count

def main():
    """
    The main function that calls the other functions and prints the answer.
    """
    lst = ['D', '9987', '4565', 'j', 'V', 'S', '1033', '7319', 'S']
    count = count_numerical_elements(lst)
    print(count)

main()

```
\end{lstlisting}

\subsection{Models}
\label{sec:appendix_model}
Table~\ref{tab:exp_models} lists the models involved in our experiments, including their names, versions, and corresponding URL links. 
We follow the licences (which can be found in the URL links) of these models to use them.
For the open-source models, we use the model weights provided by Huggingface\footnote{\href{https://huggingface.co/models}{https://huggingface.co/models}}.

\input{tables/table_models}

\subsection{Zero-Shot Prompts}
\label{sec:appendix_zero_shot}
In the zero-shot setting, we directly use the test question as the input to the models. 
For our models with CoT/AoT-finetuning, we are able to extract the predicted answers from the responses in a fixed format. 
For open-source instruction finetuned models (e.g., Llama-3-8B-Instruct), we utilize a simple yet effective instruction to guide the model to output the answer in a fixed format. 
The instruction we use is: ``\texttt{Answer the question and put the final answer in \textbackslash boxed\{\}.}''

\subsection{Few-Shot Prompts}
\label{sec:appendix_few_shot}
In the few-shot setting, we conduct experiments with two type of prompts: CoT prompt and AoT prompt.
The CoT prompt is collected from the official GitHub repository of BBH\footnote{\href{https://github.com/suzgunmirac/BIG-Bench-Hard/tree/main/cot-prompts}{https://github.com/suzgunmirac/BIG-Bench-Hard/tree/main/cot-prompts}}.
For each test task, the CoT prompt contains three demonstration questions along with their CoT rationales.
For the AoT prompt, we employ the same questions and manually create the reasoning processes in AoT format. 
In consistency with the \ourdataset, we use the Python programming language to express the reasoning process for some tasks.

For the following tasks we use a programming language reasoning process: \textit{Boolean expressions, Dyck languages, Geometric shapes, Logical deduction five objects, Logical deduction seven objects, Logical deduction three objects, Multistep arithmetic two, Navigate, Object counting, Penguins in a table, Temporal sequences, Tracking shuffled objects five objects, Tracking shuffled objects seven objects, Tracking shuffled objects three objects, Web of lies, Word sorting.}

For the following tasks we use a natural language reasoning process: 
\textit{Causal judgement, Date understanding, Disambiguation qa, Formal fallacies, Hyperbaton, Movie recommendation, Reasoning about colored objects, Ruin names, Salient translation error detection, Snarks, Sports understanding.}

Listing~\ref{lst:prompt_bbh_aot} presents examples of the Few-shot AoT prompts for BBH. 
Considering that the full prompts for all tasks in BBH is long, we show two demonstrations in the prompts here for brevity (one in natural language and one in programming language).
We will release the prompt we used after publication.

\begin{lstlisting}[caption={Examples of Few-shot AoT Prompts}, label={lst:prompt_bbh_aot}]
====== AoT Prompt for Sports Understanding =====

** Demonstration Input **
Is the following sentence plausible? "DeMar DeRozan was called for the goal tend."

** Demonstration Output **
Step 1: Recall the infomation of the sport player.
DeMar DeRozan is an American basketball player.

Step 2: Recall the infomation of the sport.
Goal tending is part of basketball.

Step 3: Evaluate the plausibility of the sentence.
Given that DeMar DeRozan is a basketball player, it is plausible that he could be called for a goal tend during a game.

Therefore, the answer is \boxed{yes}.

========== AoT Prompt for Web of Lies ==========

** Demonstration Input **
Question: Kristian tells the truth. Michaela says Kristian lies. Raymond says Michaela tells the truth. Osvaldo says Raymond tells the truth. Jamey says Osvaldo tells the truth. Does Jamey tell the truth?

** Demonstration Output **
```python
def evaluate_statements(initial_condition, statements):
    """
    This function takes a dictionary of statements where each key is a person and the value is the person they are 
    commenting on and whether they believe that person is telling the truth or not. 
    """
    truth_values = initial_condition
    for person, (target, statement) in statements.items():
        if statement == 'truth':
            truth_values[person] = truth_values[target]
        else:
            truth_values[person] = not truth_values[target]
    return truth_values

def main():
    """
    The main function that calls the other functions and prints the answer.
    """
    initial_condition = {'Kristian': True}
    statements = {
        'Michaela': ('Kristian', 'lies'),
        'Raymond': ('Michaela', 'truth'),
        'Osvaldo': ('Raymond', 'truth'),
        'Jamey': ('Osvaldo', 'truth')
    }
    truth_values = evaluate_statements(initial_condition, statements)
    print('Yes' if truth_values['Jamey'] else 'No')

main()
```
\end{lstlisting}

\subsection{AoT2CoT}
\label{sec:appendix_aot2cot}
To construct AoT2CoT for ablation study, we use LLM to transform the reasoning process from AoT format to CoT format.
We sample 10k of data from the \ourdataset (5k from AoT-Text and 5k from AoT-Code).
Our aim is to transform the reasoning format while keeping the reasoning content is roughly the same.
We prompt GPT-3.5-Turbo (the same back-end LLM for collecting \ourdataset) to transform the input reasoning process into CoT format.
In order to avoid the model generating new reasoning content in response to the question, we do not input the question and answer, but only the reasoning process.
Listing~\ref{lst:prompt_aot2cot} gives the specific prompt we use.

\begin{lstlisting}[caption={Prompts for constructing AoT2CoT. We show the system prompt and one of the three demonstrations.}, label={lst:prompt_aot2cot}]
Prompt for Converting AoT to CoT with Natural Language

** System Prompt ** 
You are a text rewriter. Given a step-by-step reasoning process, you need to reorganize its contents to meet the following requirements: 
- The rewritten text do not add new content and the reasoning logic in it is consistent with the given text. 
- The rewritten text should be as concise and natural as possible, without any explicit steps or structures.

** Demonstration Input ** 
Step 1: Extract key information about touchdowns from the text.
The touchdowns mentioned are:
- A 1-yard TD pass to Calvin Johnson
- A 7-yard TD pass to Donald Lee
- A 7-yard catch by Donald Driver
- A 21-yard catch by James Jones
- A 38-yard interception return for a touchdown by Charles Woodson

Step 2: Sort the yardages in descending order.
38 yards, 21 yards, 7 yards, 7 yards, 1 yard

Step 3: Identify the second longest touchdown.
The second longest touchdown is 21 yards.

Therefore, the answer is \boxed{21}.

** Demonstration Output ** 
The text mentions several touchdowns: a 1-yard TD pass to Calvin Johnson, a 7-yard TD pass to Donald Lee, a 7-yard catch by Donald Driver, a 21-yard catch by James Jones, and a 38-yard interception return for a touchdown by Charles Woodson. When these yardages are sorted in descending order, we get 38 yards, 21 yards, 7 yards, 7 yards, and 1 yard. The second longest touchdown is 21 yards. Therefore, the answer is \boxed{21}.

Prompt for Converting AoT to CoT with Programming Language

** System Prompt ** 
You are a Python code refactoring tool. Given a piece of Python code, you need to refactor it to meet the following requirements: 
- The refactored code should have the same functionality and `print` output as the given code. 
- The refactored code should not contain any function or classes.

** Demonstration Input ** 
```python
def count_word_frequency(sentence, word):
    """
    Count the frequency of a given word in a sentence.
    """
    words = sentence.split()
    count = words.count(word)
    return count

def compare_word_frequency(sentence1, sentence2, word):
    """
    Compare the frequency of a given word in two sentences.
    """
    count1 = count_word_frequency(sentence1, word)
    count2 = count_word_frequency(sentence2, word)
    return count1 == count2

def main():
    """
    The main function that calls the other functions and prints the answer.
    """
    sentence1 = 'a little boy that is holding a skateboard'
    sentence2 = 'this photo displays a messy desk that has two laptops on it among many other things'
    word = 'a'
    is_equal = compare_word_frequency(sentence1, sentence2, word)
    print('Yes' if is_equal else 'No')

main()
```

** Demonstration Output ** 
```python
sentence1 = 'a little boy that is holding a skateboard'
sentence2 = 'this photo displays a messy desk that has two laptops on it among many other things'
word = 'a'

is_equal = sentence1.split().count(word) == sentence2.split().count(word)
print('Yes' if is_equal else 'No’)
```
\end{lstlisting}

\subsection{\revise{\ourcot}}

\revise{\ourcot and \ourdataset utilize the same back-end LLM for data generation and the same hybrid reasoning strategy. 
Unlike AoT2CoT, we directly prompt GPT-3.5-Turbo to generate CoT responses. 
Listing~\ref{lst:prompt_cot_direct} provides the prompt we use. We modify the prompt used for generating AoT responses (Listing~\ref{lst:prompt_aot_collect}) by removing the constraints of the AoT format, thereby guiding the model to generate CoT responses. 
We then apply the same filtering approach before using the data for fine-tuning.
It is worth noting that, for the same question, the reasoning process in \ourcot may differ from the corresponding AoT reasoning process not only in the reasoning format, but also in the content of the reasoning.
This is because we cannot guarantee that the back-end LLM would generate the same reasoning content for the same question in two independent generations.}

\begin{lstlisting}[caption={Prompts for constructing \ourcot. We show the system prompt and one of the three demonstrations.}, label={lst:prompt_cot_direct}]
Prompt for Collecting CoT (Direct) Responses with Natural Language

** System Prompt **
You are a powerful reasoning model. I will present you with a question and the correct answer. Your task is to output the intermediate reasoning process step-by-step and conclude the reasoning process by stating the final answer as "The answer is \boxed{answer}".

** Demonstration Input **
Which sentence has the correct adjective order: Options:
(A) Indonesian triangular obnoxious gold red shoe
(B) obnoxious triangular red Indonesian gold shoe
Correct Answer: (B)

** Demonstration Output **
In English, adjectives generally follow this order: quantity, quality, size, age, shape, color, origin, material, purpose. We find that option (A) has the adjectives in the wrong order. On the other hand, option (B) correctly orders the adjectives as quality, shape, color, origin, material. 

Therefore, the answer is \boxed{(B)}.

================================================
Prompt for Collecting CoT (Direct) Responses with Programming Language

** System Prompt **
You are a powerful reasoning model and Python coder. I will present you with a question and the correct answer. Your task is to output a Python program to answer the question and print the correct answer.

** Demonstration Input **
In this task, answer 'Yes' if the frequency of the given word in the two sentences is equal, else answer 'No’. Sentence1: 'a little boy that is holding a skateboard', Sentence2: 'this photo displays a messy desk that has two laptops on it among many other things'. Is the frequency of the word 'a' in two sentences equal?
Correct Answer: No

** Demonstration Output **
```python
sentence1 = 'a little boy that is holding a skateboard'
sentence2 = 'this photo displays a messy desk that has two laptops on it among many other things'
word = 'a'
is_equal = sentence1.split().count(word) == sentence2.split().count(word)
print('Yes' if is_equal else 'No')
```
\end{lstlisting}

\subsection{BBH Benchmark}
\label{sec:appendix_bbh}
\input{tables/table_bbh}

Table~\ref{tab:bbh} presents the tasks and number of questions within the BBH benchmark.
Following~\citet{DBLP:conf/acl/SuzgunSSGTCCLCZ23}, tasks are divided into two categories: NLP tasks and algorithm tasks.
Tasks ``Logical deduction'' and ``Tracking shuffled objects'' consist of 3 sub-tasks.
The questions and answers in BBH are in English.

\section{Additional Experiment Results}
\label{sec:appendix_add_result}

\subsection{Performances on Subtasks}
\label{sec:appendix_by_tasks}
Table~\ref{tab:bbh_by_tasks} shows the performance on subtasks of BBH.
We also introduce human performance as a reference~\cite{DBLP:conf/acl/SuzgunSSGTCCLCZ23}.
We can observe that AoT-finetuning compared to cot-finetuning is able to achieve improvements on multiple reasoning tasks on multiple models.

\subsection{Few-Shot Performance}
\label{sec:appendix_few_details}

Table~\ref{tab:few-shot-detail} presents the 3-shot performance on BBH, including those that are instruction-finetuned (such as Llama-3-8B-Instruct). 
A consistent trend can be observed where instruction-finetuned models achieve lower few-shot performance compared to models that are merely pre-trained.
For example, Llama-3-8B-Instruct achieves an overall accuracy of 56.7\% with the official few-shot CoT prompt, while Llama-3-8B can achieve 60.0\% with the same prompt.
For the CodeLlama-7B-Instruct and Llama-2-7B-Chat, a significant performance decrease is noted in the instruction-finetuned versions. 
Upon examining their responses, we find that in most cases, the models did not follow the demonstration format for answering. 
For the Llama-3-8B-Instruct and Mistral-7B-Instruct, the responses adhered to the demonstration format. 
In this scenario, using the AoT prompt yields better results than the CoT prompt.

\input{tables/table_few_details}
\input{tables/table_by_tasks}

\input{tables/table_error}

\subsection{Case Study}
\label{sec:appendix_case_study}

Listing~\ref{lst:success_cases} shows success cases for the AoT-finetuned Llama-3-8B.
The AoT-finetuned model demonstrates a certain degree of ability to reason with abstraction.
For example, for the Case 3, the model first defines the tools needed to solve the problem at an abstract level, i.e., the \texttt{Person} class and the \texttt{swap\_gifts} function.
Subsequently, the model then utilizes these tools to solve the problem based on the concrete conditions of the question.
Listing~\ref{lst:fail_cases} presents the error cases.

\subsection{Computational Cost Comparison}
\label{sec:appendix_computational_cost}

Considering that our AoT format encompasses multi-level abstractions, the average length of AoT responses tends to be longer than that of CoT responses. 
Consequently, the AoT-finetuned model may takes a higher computational cost during inference to generate more tokens. 
To clarify the impact of this in practical applications, we conduct a brief comparison of computational costs. 
We calculate the average length of responses from Llama-3-8B-CoT and Llama-3-8B-AoT on the BBH questions. 
On average, the responses of Llama-3-8B-CoT consist of 107.1 tokens, while the responses of Llama-3-8B-AoT consist of 188.9 tokens.
Thus, AoT responses takes approximately 80\% more tokens than CoT responses. 

To estimate the generation time during inference, we use the vllm\footnote{\href{https://github.com/vllm-project/vllm}{https://github.com/vllm-project/vllm}} library on a server with 8 A100-SXM4-40GB GPUs, setting the batch size to 1. 
The average time for Llama-3-8B-AoT to generate one AoT response is 1.96 seconds. 
However, this estimation should only be taken as a rough reference, as it may be influenced by various factors such as hardware conditions.

\onecolumn

\begin{lstlisting}[caption={Success cases of Llama-3-8B-AoT.},backgroundcolor=\color{green!10!white}, label={lst:success_cases}]
==================================== Case 1 from Task Hyperbaton ====================================

**Input**
Question:Which sentence has the correct adjective order:
Options:
(A) lovely tiny old-fashioned pyramidal Pakistani glass eating cat
(B) pyramidal lovely old-fashioned Pakistani tiny eating glass cat

**AoT Prediction**
Step 1: Identify the correct adjective order.
The correct adjective order is: descriptive adjective + size adjective + age adjective + origin adjective + material adjective + function adjective + animal adjective.

Step 2: Analyze each option.
(A) lovely tiny old-fashioned pyramidal Pakistani glass eating cat
(B) pyramidal lovely old-fashioned Pakistani tiny eating glass cat

Step 3: Compare the options to the correct adjective order.
Option (A) follows the correct adjective order: lovely (descriptive) + tiny (size) + old-fashioned (age) + pyramidal (shape) + Pakistani (origin) + glass (material) + eating (function) + cat (animal).
Option (B) does not follow the correct adjective order: pyramidal (shape) + lovely (descriptive) + old-fashioned (age) + Pakistani (origin) + tiny (size) + eating (function) + glass (material) + cat (animal).

Therefore, the answer is \boxed{(A)}.

=============================== Case 2 from Task Movie Recommendation ===============================

**Input**
Question: Find a movie similar to Independence Day, Star Wars Episode V - The Empire Strikes Back, American Beauty, Star Wars Episode VI - Return of the Jedi:
Options:
(A) Frankenweenie
(B) Ivan Vasilievich Back to the Future
(C) We Bought a Zoo
(D) Star Wars Episode IV - A New Hope

**AoT Prediction**
Step 1: Analyze the given movies.
- Independence Day: A science fiction movie about aliens invading Earth.
- Star Wars Episode V - The Empire Strikes Back: A science fiction movie about the ongoing conflict between the Rebel Alliance and the Galactic Empire.
- American Beauty: A drama movie about a suburban family and their struggles with life.
- Star Wars Episode VI - Return of the Jedi: A science fiction movie about the final battle between the Rebel Alliance and the Galactic Empire.

Step 2: Identify the common theme or genre.
The common theme among the given movies is science fiction.

Step 3: Compare the options.
(A) Frankenweenie: A stop-motion animated film about a boy and his dog, not a science fiction movie.
(B) Ivan Vasilievich Back to the Future: A Russian comedy film about a man who accidentally travels back in time, not a science fiction movie.
(C) We Bought a Zoo: A drama movie about a family buying a zoo, not a science fiction movie.
(D) Star Wars Episode IV - A New Hope: A science fiction movie set in a galaxy far, far away, similar to the given movies.

Step 4: Select the correct answer.
The movie that is similar to Independence Day, Star Wars Episode V - The Empire Strikes Back, American Beauty, and Star Wars Episode VI - Return of the Jedi is Star Wars Episode IV - A New Hope.

Therefore, the answer is \boxed{D}.

============================ Case 3 from Task Tracking shuffled Objects ============================

**Input**
Question: Alice, Bob, Claire, Dave, and Eve are holding a white elephant gift exchange. At the start of the event, they are each holding a present of a different color: Alice has a brown present, Bob has a black ball, Claire has a blue present, Dave has a orange ball, and Eve has a red present. As the event progresses, pairs of people swap gifts. First, Claire and Alice swap their gifts. Then, Bob and Eve swap their gifts. Then, Dave and Bob swap their gifts. Then, Alice and Claire swap their gifts. Finally, Eve and Dave swap their gifts. At the end of the event, Claire has the
Options: (A) brown present (B) black ball (C) blue present (D) orange ball (E) red present

**AoT Prediction**
```python
class Person:
    def __init__(self, name, gift):
        self.name = name
        self.gift = gift

def swap_gifts(person1, person2):
    """
    Swap the gifts between two people.
    """
    temp = person1.gift
    person1.gift = person2.gift
    person2.gift = temp

def main():
    """
    The main function that calls the other functions and prints the answer.
    """
    alice = Person('Alice', 'brown')
    bob = Person('Bob', 'black')
    claire = Person('Claire', 'blue')
    dave = Person('Dave', 'orange')
    eve = Person('Eve', 'red')

    swap_gifts(claire, alice)
    swap_gifts(bob, eve)
    swap_gifts(dave, bob)
    swap_gifts(alice, claire)
    swap_gifts(eve, dave)

    if claire.gift == 'brown':
        print('A')
    elif claire.gift == 'black':
        print('B')
    elif claire.gift == 'blue':
        print('C')
    elif claire.gift == 'orange':
        print('D')
    elif claire.gift == 'red':
        print('E')

main()
```

=============================== Case 4 from Task Penguins in a Table ===============================

**Input**
Question: Here is a table where the first line is a header and each subsequent line is a penguin: name, age, height (cm), weight (kg) Louis, 7, 50, 11 Bernard, 5, 80, 13 Vincent, 9, 60, 11 Gwen, 8, 70, 15 For example: the age of Louis is 7, the weight of Gwen is 15 kg, the height of Bernard is 80 cm. What is the name of the last penguin sorted by alphabetic order?
Options: (A) Louis (B) Bernard (C) Vincent (D) Gwen (E) James

**AoT Prediction**
```python
def sort_penguins(penguins):
    """
    Sort a list of penguins by their names in alphabetical order.
    """
    sorted_penguins = sorted(penguins, key=lambda penguin: penguin[0])
    return sorted_penguins

def get_last_penguin(sorted_penguins):
    """
    Get the last penguin in the sorted list.
    """
    last_penguin = sorted_penguins[-1]
    return last_penguin[0]

def main():
    """
    The main function that calls the other functions and prints the answer.
    """
    penguins = [['Louis', 7, 50, 11], ['Bernard', 5, 80, 13], ['Vincent', 9, 60, 11], ['Gwen', 8, 70, 15]]
    sorted_penguins = sort_penguins(penguins)
    last_penguin = get_last_penguin(sorted_penguins)
    options = ['Louis', 'Bernard', 'Vincent', 'Gwen', 'James']
    answer = options.index(last_penguin) + 1
    print(chr(96 + answer))

main()
```
\end{lstlisting}

\begin{lstlisting}[caption={Error cases of Llama-3-8B-AoT.},backgroundcolor=\color{blue!10!white},label={lst:fail_cases}]
==================================== Error Case: Reasoning Error ====================================

**Input**
Question: Rashida lies. Osvaldo says Rashida tells the truth. Sherrie says Osvaldo tells the truth. Gwenn says Sherrie tells the truth. Jerry says Gwenn lies. Does Jerry tell the truth?

**AoT Prediction**
Step 1: Analyze the statements made by each person.
- Rashida lies.
- Osvaldo says Rashida tells the truth.
- Sherrie says Osvaldo tells the truth.
- Gwenn says Sherrie tells the truth.
- Jerry says Gwenn lies.

Step 2: Evaluate the statements based on the given information.
- Rashida lies, so her statement is not reliable.
- Osvaldo says Rashida tells the truth, but since Rashida lies, Osvaldo's statement is not reliable.
- Sherrie says Osvaldo tells the truth, but since Osvaldo's statement is not reliable, Sherrie's statement is not reliable.
- Gwenn says Sherrie tells the truth, but since Sherrie's statement is not reliable, Gwenn's statement is not reliable.
- @\hlerror{Jerry says Gwenn lies, but since Gwenn's statement is not reliable, Jerry's statement is not }@
@\hlerror{reliable.}@

Step 3: Determine if Jerry tells the truth.
Based on the given information, Jerry's statement is not reliable.

Therefore, the answer is \boxed{no}.

**Predicion**
no
**Ground Truth**
Yes
**Error Explanation**
The model derives incorrect logical conclusions from given conditions.

=================================== Error Case: Knowledge Lacking ===================================

**Input**
Is the following sentence plausible? "Courtland Sutton hit a triple."

**AoT Prediction**
Step 1: Analyze the sentence structure and meaning.
The sentence states that "Courtland Sutton hit a triple."

Step 2: Recall the definition of a triple in baseball.
In baseball, a triple is a hit in which the batter reaches third base safely after hitting the ball.

Step 3: Evaluate the plausibility of the sentence.
@\hlerror{It is plausible for Courtland Sutton to hit a triple in baseball, as it is a possible outcome for a}@
@\hlerror{batter.}@

Therefore, the answer is \boxed{Yes}.

**Predicion**
Yes
**Ground Truth**
No
**Error Explanation**
The model lacks the necessary knowledge that Courtland Sutton usually refers to the American football wide receiver for the Denver Broncos of the National Football League (NFL). 

================================= Error Case: Task Misunderstanding =================================

**Input**
True or False or not True or False is

**AoT Prediction**
Step 1: Analyze the given statement.
The statement is "True or False or not True or False is."

Step 2: Evaluate the statement.
@\hlerror{The statement is grammatically incorrect and does not make sense. It is not clear what the statement}@
@\hlerror{is trying to convey.}@

Step 3: Determine the correct answer.
Since the statement is not true or false, the correct answer is \boxed{not True or False}.

**Predicion**
not True or False
**Ground Truth**
True
**Error Explanation**
The model fails to understand that the task is to simplify the input boolean expression. 

================================= Error Case: Condition Repetition ==================================

**Input**
Which of the following is a humorous edit of this artist or movie name: 'guns n' roses'?
Options:
(A) guns n' goses
(B) suns n' roses
(C) guns n' rosjs
(D) guns n' rgoses

**AoT Prediction**
Step 1: Analyze the given options.
(A) guns n' goses
(B) suns n' roses
(C) guns n' rosjs
(D) guns n' rgoses

Step 2: Identify the humorous edit.
@\hlerror{The humorous edit of "guns n' roses" is "guns n' goses".}@

Therefore, the answer is \boxed{(A)}.

**Predicion**
(A)
**Ground Truth**
(B)
**Error Explanation**
The reasoning process simply repeats the given conditions and then leap directly to the predicted answer.

===================================== Error Case: Hallucination =====================================

**Input**
I have two garlics, a cabbage, a stalk of celery, five carrots, a lettuce head, four heads of broccoli, a potato, an onion, and a yam. How many vegetables do I have?

**AoT Prediction**
```python
def count_vegetables(vegetables):
    """
    Count the number of vegetables in a list.
    """
    count = len(vegetables)
    return count

def main():
    """
    The main function that calls the other functions and prints the answer.
    """
@\hlerror{\qquad vegetables = ['garlic', 'cabbage', 'celery', 'carrot', 'carrot', 'carrot', 'carrot', 'carrot',}@
@\hlerror{'carrot', 'lettuce', 'broccoli','broccoli', 'broccoli', 'broccoli', 'broccoli', 'broccoli', 'potato',}@
@\hlerror{'onion', 'yam']}@
    count = count_vegetables(vegetables)
    print(count)

main()
```

**Predicion**
19
**Ground Truth**
17
**Error Explanation**
The model fabricates non-existent conditions or misinterprets given conditions. For example, the input states that there are 5 carrots, but 6 carrots are included in the variable.

===================================== Error Case: Runtime Error =====================================

**Input**
Here is a table where the first line is a header and each subsequent line is a penguin: name, age, height (cm), weight (kg) Louis, 7, 50, 11 Bernard, 5, 80, 13 Vincent, 9, 60, 11 Gwen, 8, 70, 15 For example: the age of Louis is 7, the weight of Gwen is 15 kg, the height of Bernard is 80 cm. How many penguins are more than 8 years old?
Options:
(A) 1
(B) 2
(C) 3
(D) 4
(E) 5

**AoT Prediction**
```python
def count_penguins_over_age(penguins, age):
    """
    Count the number of penguins that are older than a given age.
    """
    count = sum(1 for penguin in penguins if int(penguin[1]) > age)
    return count

def main():
    """
    The main function that calls the other functions and prints the answer.
    """
@\hlerror{\qquad penguins = ['Louis, 7, 50, 11', 'Bernard, 5, 80, 13', 'Vincent, 9, 60, 11', 'Gwen, 8, 70, 15']}@
    age = 8
    count = count_penguins_over_age(penguins, age)
    options = {1: 'A', 2: 'B', 3: 'C', 4: 'D', 5: 'E'}
    answer = options.get(count, 'Option not found')
    print(answer)

main()
```

**Predicion**
None
**Ground Truth**
(A)
**Error Explanation**
The generated program is inexecutable.

===================================== Error Case: False Negative ====================================

**Input**
Is the following sentence plausible? "Andres Iniesta performed a give and go."

**AoT Prediction**
Step 1: Understand the meaning of the sentence.
The sentence "Andres Iniesta performed a give and go" suggests that Andres Iniesta participated in a specific football technique called a "give and go".

Step 2: Recall the definition of a "give and go" in football.
A "give and go" is a football technique where a player passes the ball to a teammate, then quickly moves towards the space created by the teammate's run. The teammate then returns the ball to the player who initiated the move.

Step 3: Evaluate the plausibility of the sentence.
Given that Andres Iniesta is a professional football player, it is plausible that he would perform a "give and go" technique during a game.

Therefore, the sentence "Andres Iniesta performed a give and go" is \boxed{plausible}.

**Predicion**
plausible
**Ground Truth**
Yes
**Error Explanation**
The correct predicted answer does not fully match the ground truth.
\end{lstlisting}

\input{tables/table_aot_task_detail}

\end{document}

%% file: tables/table_zero_shot.tex
\begin{table}[t]
\centering
\small
\begin{tabular}{@{}l|ccc@{}}
\toprule
\textbf{Model} & \textbf{NLP} & \textbf{Alg} & \textbf{All} \\
\midrule
\multicolumn{4}{c}{\textit{Proprietary Models}} \\ \midrule
GPT-3.5-Turbo & 48.2 & 39.2 & 43.9 \\
GPT-4 & \textbf{71.4} & \textbf{59.5} & \textbf{65.7} \\ \midrule
\multicolumn{4}{c}{\textit{Instruction-Finetuned Models}} \\ \midrule
Llama-3-8B-Instruct & \textbf{56.2} & \textbf{41.0} & \textbf{49.0} \\
CodeLlama-7B-Instruct & 34.4 & 23.2 & 29.1 \\
Llama-2-7B-Chat & 31.0 & 17.4 & 24.5 \\
Mistral-7B-Instruct & 34.3 & 21.8 & 28.3 \\ \midrule
\multicolumn{4}{c}{\textit{Our Finetuned Models}} \\ \midrule
Llama-3-8B-CoT & 51.4 & 35.2 & 43.6 \\
Llama-3-8B-AoT & \textbf{51.7} & \textbf{55.0} & \textbf{53.3} {\scriptsize \textbf{(+9.7)}}\\ \midrule
CodeLlama-7B-CoT & 47.0 & 31.6 & 39.6 \\
CodeLlama-7B-AoT & \textbf{49.8} & \textbf{49.3} & \textbf{49.6} {\scriptsize \textbf{(+10.0)}} \\ \midrule
Llama-2-7B-CoT & \textbf{42.1} & 22.5 & 32.8 \\
Llama-2-7B-AoT & 41.1 & \textbf{29.0} & \textbf{35.4} {\scriptsize \textbf{(+2.6)}}\\ \midrule
Mistral-7B-CoT & 53.8 & 34.3 & 44.5 \\
Mistral-7B-AoT & \textbf{55.0} & \textbf{48.4} & \textbf{51.8} {\scriptsize \textbf{(+7.3)}} \\
\bottomrule
\end{tabular}
\caption{Evaluation performance (\%) on the unseen BBH benchmark under the \textbf{zero-shot} setting (realistic setting). 
$X$-AoT/CoT indicates the language model $X$ with AoT/CoT-finetuning.
\revise{AoT/CoT-finetuning shares the same training questions and training data scale, but differs in the reasoning processes.}
We mark the improvements of AoT over CoT in parentheses.
}
\label{tab:Zero-shot}
\vspace{-5mm}
\end{table}

%% file: tables/table_few_shot.tex
\begin{table}[t]
\centering
\small
\resizebox{\columnwidth}{!}{%
\begin{tabular}{@{}lc|ccc@{}}
\toprule
\textbf{Model} & \textbf{Prompt} & \textbf{NLP} & \textbf{Alg} & \textbf{All} \\
\midrule
\multicolumn{5}{c}{\textit{Proprietary Models}} \\ \midrule
GPT-3.5-Turbo & CoT & 69.0 & 70.2 & 69.6 \\
GPT-3.5-Turbo & AoT & \textbf{72.1} & \textbf{94.3} & \textbf{82.7} {\scriptsize \textbf{(+13.1)}} \\ \midrule
GPT-4 & CoT & \textbf{87.0} & 86.4 & 86.7 \\
GPT-4 & AoT & 86.7 & \textbf{97.0} & \textbf{91.6} {\scriptsize \textbf{(+4.9)}} \\ \midrule
\multicolumn{5}{c}{\textit{Pre-Trained Models}} \\ \midrule
Llama-3-8B & CoT & 63.6 & 56.2 & 60.0 \\
Llama-3-8B & AoT & \textbf{68.1} & \textbf{82.8} & \textbf{75.1} {\scriptsize \textbf{(+15.1)}} \\ \midrule
CodeLlama-7B & CoT & 48.8 & 33.4 & 41.4 \\
CodeLlama-7B & AoT & \textbf{52.1} & \textbf{81.1} & \textbf{66.0} {\scriptsize \textbf{(+24.6)}} \\ \midrule
Llama-2-7B & CoT & 47.2 & 27.6 & 37.8 \\
Llama-2-7B & AoT & \textbf{47.5} & \textbf{66.3} & \textbf{56.5} {\scriptsize \textbf{(+18.7)}} \\ \midrule
Mistral-7B & CoT & 61.1 & 48.9 & 55.2 \\
Mistral-7B & AoT & \textbf{62.3} & \textbf{80.4} & \textbf{71.0} {\scriptsize \textbf{(+15.8)}} \\ \midrule
\multicolumn{5}{c}{\textit{Our Finetuned Models}} \\ \midrule
Llama-3-8B-CoT & CoT & 51.0 & 36.2 & 43.9 \\
Llama-3-8B-AoT & AoT & \textbf{73.1} & \textbf{92.8} & \textbf{82.5} {\scriptsize \textbf{(+38.6)}} \\ \midrule
CodeLlama-7B-CoT & CoT & 47.3 & 32.6 & 40.3 \\
CodeLlama-7B-AoT & AoT & \textbf{55.4} & \textbf{85.6} & \textbf{69.9} {\scriptsize \textbf{(+29.6)}} \\ \midrule

Llama-2-7B-CoT & CoT & 51.1 & 32.0 & 41.9 \\
Llama-2-7B-AoT & AoT & \textbf{55.4} & \textbf{75.2} & \textbf{64.9} {\scriptsize \textbf{(+23.0)}} \\ \midrule

Mistral-7B-CoT & CoT & 61.4 & 40.8 & 51.6 \\
Mistral-7B-AoT & AoT & \textbf{71.3} & \textbf{84.8} & \textbf{77.7} {\scriptsize \textbf{(+26.1)}} \\

\bottomrule
\end{tabular}
}
\caption{Evaluation performance (\%) on BBH with \textbf{3-shot} prompting (standard setting). 
We prompt models with 3 demonstrations in CoT/AoT reasoning format.
}
\label{tab:Few-shot}
\vspace{-5mm}
\end{table}

%% file: tables/table_aot2cot.tex
\begin{table}[t]
\small
\centering
\resizebox{\columnwidth}{!}{%
\begin{tabular}{@{}llccc@{}}
\toprule
\textbf{Model}                      & \textbf{Training Data Format}  & \textbf{NLP}  & \textbf{Alg}  & \textbf{All}  \\ \midrule
\multirow{4}{*}{Llama-3-8B} & CoT (CoT Collection)     & 46.9 & 32.0 & 39.7 \\ \cmidrule(l){2-5}
                            & \ourcot*  & 47.6 & 45.2 & 46.5 \\
                            & AoT2CoT * & 46.3 & 39.4 & 43.0 \\
                            & AoT (\ourdataset) *    & \textbf{48.9} & \textbf{50.6} & \textbf{49.7} \\ \midrule
\multirow{4}{*}{Mistral-7B} & CoT (CoT Collection)     & 45.9 & 30.3 & 38.4 \\ \cmidrule(l){2-5}
                            & \ourcot*  & 46.3 & 45.6 & 46.0 \\ 
                            & AoT2CoT * & 44.4 & 47.9 & 46.1 \\
                            & AoT (\ourdataset) *     & \textbf{48.3} & \textbf{52.1} & \textbf{50.1} \\ \bottomrule
\end{tabular}%
}
\caption{Ablation on the reasoning format of training data. We finetune models with 10k questions with reasoning processes in different format and report their zero-shot performance.
\revise{Data with * are collected with the same back-end LLM and hybrid reasoning strategy.}}
\label{tab:aot2cot}
\vspace{-5mm}
\end{table}

%% file: tables/table_model_analysis.tex
\begin{table}[t]
\setlength{\tabcolsep}{3pt}
\small
\centering
\resizebox{\columnwidth}{!}{%
\begin{tabular}{@{}lcccccc@{}}
\toprule
& \multicolumn{3}{c}{\textbf{Text}} & \multicolumn{3}{c}{\textbf{Code}} \\ 
\cmidrule(lr){2-4} \cmidrule(lr){5-7}
\textbf{Model} & \textbf{UR} & \textbf{FC} & \textbf{AC} & \textbf{UR} & \textbf{FC} & \textbf{AC} \\
\midrule
Llama-3-8B-AoT & 73.4 & 99.3 & 50.3 & 26.6 & 85.1 & 53.7 \\
CodeLlama-7B-AoT & 84.9 & 99.8 & 43.6 & 15.1 & 96.9 & 61.1 \\
Llama-2-7B-AoT & 90.3 & 96.1 & 34.6 & 9.7 & 56.9 & 24.9 \\
Mistral-7B-AoT & 84.7 & 99.7 & 49.1 & 15.3 & 87.3 & 50.4 \\
\bottomrule
\end{tabular}
}
\caption{Response analysis of AoT-finetuned models.
UR=Usage Rate. FC=Format Correctness. AC=Answer Correctness.
}
\label{tab:ablation_model_analysis}
\vspace{-5mm}
\end{table}

%% file: tables/table_aot_task_counter.tex
{
\small

Question Answering (65); \par
Text Categorization (24); \par
Question Understanding (11); \par
Multiple-Choice QA  (11); \par
Information Extraction (10); \par
Dialogue Generation (10); \par
Toxic Language Detection (10); \par
Text Matching (9); \par
Answerability Classification (9); \par
Textual Entailment (9); \par
Sentiment Analysis (9); \par
Natural Language Inference (8); \par
Speaker Identification (7); \par
Inverted Natural Language Inference (7); \par
Fill in The Blank (7); \par
Sentence Composition (7); \par
Closed-Book QA (6); \par
Text Quality Evaluation (6); \par
Inverted Multiple-Choice QA  (6); \par
Extractive QA (6); \par
Intent Identification (5); \par
Text Completion (5); \par
Explanation (5); \par
Story Composition (4); \par
Named Entity Recognition (4); \par
Question Rewriting (4); \par
Commonsense Classification (4); \par
Dialogue Act Recognition (4); \par
Answer Verification (4); \par
Conversational Question Answering (4); \par
Inverted Extractive QA (4); \par
Answer Generation (4); \par
Coherence Classification (3); \par
Adversarial QA (3); \par
Inverted Closed-Book QA (3); \par
Program Execution (3); \par
Word Semantics (3); \par
Dialogue State Tracking (3); \par
Fact Verification (3); \par
Question Generation (3); \par
Word Relation Classification (2); \par
Linguistic Probing (2); \par
Question Decomposition (2); \par
Text to Code (2); \par
Cause Effect Classification (2); \par
Grammar Error Detection (2); \par
Stereotype Detection (2); \par
Inverted Mathematical QA (2); \par
Mathematical QA (2); \par
Summarization (2); \par
Gender Classification (2); \par
Pos Tagging (2); \par
Keyword Tagging (2); \par
Span Generation (2); \par
Sentence Ordering (2); \par
Dialog Next Turn Prediction (2); \par
Coreference Resolution (2); \par
Word Analogy (1); \par
Negotiation Strategy Detection (1); \par
Question Context Generation (1); \par
Section Classification (1); \par
Inverted Grammatical Acceptability (1); \par
Grammatical Acceptability (1); \par
Speaker Relation Classification (1); \par
Common Sense Reasoning Question Answering (1); \par
Answer Incomplete Questions (1); \par
Emotion Word Generation (1); \par
Emotional Reaction Generation (1); \par
Intent Generation (1); \par
Grade School Math Word Problems (1); \par
Entity Generation (1); \par
Entity Relation Classification (1); \par
Food Classification (1); \par
Language Identification (1); \par
Paraphrasing (1); \par
Poem Generation (1); \par
Ethics Classification (1); \par
Irony Detection (1); \par
Stance Detection (1); \par
Clock Format Conversion (1); \par
Spelling Error Detection (1); \par
Rhyme Generation (1); \par
Date Validity Prediction (1); \par
Mathematics (1); \par
Temporal Reasoning (1); \par
Leap Year Prediction (1); \par
Edible Prediction (1); \par
Inverted Coreference Resolution (1); \par
Sentence Perturbation (1); \par
}

%% file: tables/table_models.tex
\begin{table*}[t]
\centering
\resizebox{\textwidth}{!}{%
\begin{tabular}{lll}
\toprule
\textbf{Model}   & \textbf{Version}                         & \textbf{URL}                                                      \\
\midrule
Llama2-7B~\cite{DBLP:journals/corr/abs-2307-09288}     & meta-llama/Llama-2-7b-hf   & \href{https://huggingface.co/meta-llama/Llama-2-7b}{https://huggingface.co/meta-llama/Llama-2-7b}   \\
Llama2-7B-chat~\cite{DBLP:journals/corr/abs-2307-09288}     & meta-llama/Llama-2-7b-chat-hf   & \href{https://huggingface.co/meta-llama/Llama-2-7b-chat-hf}{https://huggingface.co/meta-llama/Llama-2-7b-chat-hf}   \\

\midrule
CodeLlama-7B~\cite{DBLP:journals/corr/abs-2308-12950}     & meta-llama/CodeLlama-7b-hf   & \href{https://huggingface.co/meta-llama/CodeLlama-7b-hf}{https://huggingface.co/meta-llama/CodeLlama-7b-hf}   \\
CodeLlama-7B-Instruct~\cite{DBLP:journals/corr/abs-2308-12950}     & meta-llama/CodeLlama-7b-Instruct-hf   & \href{https://huggingface.co/meta-llama/CodeLlama-7b-Instruct-hf}{https://huggingface.co/meta-llama/CodeLlama-7b-Instruct-hf}   \\

\midrule
Llama-3-8B~\cite{llama3modelcard}     & meta-llama/Meta-Llama-3-8B   & \href{https://huggingface.co/meta-llama/Meta-Llama-3-8B}{https://huggingface.co/meta-llama/Meta-Llama-3-8B}   \\
Llama-3-8B-Instruct~\cite{llama3modelcard}     & meta-llama/Meta-Llama-3-8B-Instruct   & \href{https://huggingface.co/meta-llama/Meta-Llama-3-8B-Instruct}{https://huggingface.co/meta-llama/Meta-Llama-3-8B-Instruct}   \\

\midrule
Mistral-7B~\cite{DBLP:journals/corr/abs-2310-06825}     & mistralai/Mistral-7B-v0.1   & \href{hhttps://huggingface.co/mistralai/Mistral-7B-v0.1}{https://huggingface.co/mistralai/Mistral-7B-v0.1}   \\
Mistral-7B-Instruct~\cite{DBLP:journals/corr/abs-2310-06825}     & mistralai/Mistral-7B-Instruct-v0.2   & \href{https://huggingface.co/mistralai/Mistral-7B-Instruct-v0.2}{https://huggingface.co/mistralai/Mistral-7B-Instruct-v0.2}   \\

\midrule
GPT-3.5~\cite{GPT-3.5}       & gpt-3.5-turbo-0125                   & \href{https://platform.openai.com/docs/models/gpt-3-5}{https://platform.openai.com/docs/models/gpt-3-5}        \\
GPT-4~\cite{DBLP:journals/corr/abs-2303-08774}         & gpt-4-0613                           & \href{https://platform.openai.com/docs/models/gpt-4}{https://platform.openai.com/docs/models/gpt-4}   \\    
\bottomrule
\end{tabular}
}
\caption{Detailed information about the models we experiment with.}
\vspace{+2cm}
\label{tab:exp_models}
\end{table*}

%% file: tables/table_bbh.tex
\begin{table}[t]
\small
\centering
\begin{tabular}{@{}lc@{}}
\toprule
\textbf{Task} & \textbf{\# Questions} \\
\midrule
boolean\_expressions* & 250 \\
causal\_judgement & 187 \\
date\_understanding & 250 \\
disambiguation\_qa & 250 \\
dyck\_languages* & 250 \\
formal\_fallacies & 250 \\
geometric\_shapes* & 250 \\
hyperbaton & 250 \\
logical\_deduction* & \\
\quad logical\_deduction\_five\_objects* & 250 \\
\quad logical\_deduction\_seven\_objects* & 250 \\
\quad logical\_deduction\_three\_objects* & 250 \\
movie\_recommendation & 250 \\
multistep\_arithmetic\_two* & 250 \\
navigate* & 250 \\
object\_counting* & 250 \\
penguins\_in\_a\_table & 146 \\
reasoning\_about\_colored\_objects & 250 \\
ruin\_names & 250 \\
salient\_translation\_error\_detection & 250 \\
snarks & 178 \\
sports\_understanding & 250 \\
temporal\_sequences* & 250 \\
tracking\_shuffled* & \\
\quad tracking\_shuffled\_objects\_five\_objects* & 250 \\
\quad tracking\_shuffled\_objects\_seven\_objects* & 250 \\
\quad tracking\_shuffled\_objects\_three\_objects* & 250 \\
web\_of\_lies* & 250 \\
word\_sorting* & 250 \\

\bottomrule
\end{tabular}%
\caption{Tasks in the BBH Benchmark. * indicates that the task is an algorithmic task. Untagged tasks belong to NLP tasks.}
\label{tab:bbh}
\end{table}

%% file: tables/table_few_details.tex
\begin{table}[t]
\centering
\small
\resizebox{\columnwidth}{!}{%
\begin{tabular}{@{}lcccc@{}}
\toprule
\textbf{Model} & \textbf{Prompt} & \textbf{NLP} & \textbf{Alg} & \textbf{All} \\
\midrule
Llama-3-8B & CoT & 63.6 & 56.2 & 60.0 \\
Llama-3-8B & AoT & 68.1 & 82.8 & 75.1 \\
Llama-3-8B-Instruct & CoT & 66.6 & 45.8 & 56.7 \\
Llama-3-8B-Instruct & AoT & 69.3 & 71.2 & 70.2 \\
Llama-3-8B-CoT & CoT & 51.0 & 36.2 & 43.9 \\
Llama-3-8B-AoT & AoT & 73.1 & 92.8 & 82.5 \\ \midrule
CodeLlama-7B & CoT & 48.8 & 33.4 & 41.4 \\
CodeLlama-7B & AoT & 52.1 & 81.1 & 66.0 \\
CodeLlama-7B-Instruct & CoT & 29.2 & 10.9 & 20.4 \\
CodeLlama-7B-Instruct & AoT & 29.7 & 7.3 & 19.0 \\
CodeLlama-7B-CoT & CoT & 47.3 & 32.6 & 40.3 \\
CodeLlama-7B-AoT & AoT & 55.4 & 85.6 & 69.9 \\ \midrule
Llama-2-7B & CoT & 47.2 & 27.6 & 37.8 \\
Llama-2-7B & AoT & 47.5 & 66.3 & 56.5 \\
Llama-2-7B-Chat & CoT & 37.6 & 16.5 & 27.5 \\
Llama-2-7B-Chat & AoT & 34.2 & 15.8 & 25.4 \\
Llama-2-7B-CoT & CoT & 51.1 & 32.0 & 41.9 \\
Llama-2-7B-AoT & AoT & 55.4 & 75.2 & 64.9 \\ \midrule
Mistral-7B & CoT & 61.1 & 48.9 & 55.2 \\
Mistral-7B & AoT & 62.3 & 80.4 & 71.0 \\
Mistral-7B-Instruct & CoT & 54.7 & 48.2 & 51.6 \\
Mistral-7B-Instruct & AoT & 56.1 & 75.6 & 65.4 \\
Mistral-7B-CoT & CoT & 61.4 & 40.8 & 51.6 \\
Mistral-7B-AoT & AoT & 71.3 & 84.8 & 77.7 \\ \midrule
GPT-3.5-Turbo & CoT & 69.0 & 70.2 & 69.6 \\
GPT-3.5-Turbo & AoT & 72.1 & 94.3 & 82.7 \\ \midrule
GPT-4 & CoT & 87.0 & 86.4 & 86.7 \\
GPT-4 & AoT & 86.7 & 97.0 & 91.6 \\
\bottomrule
\end{tabular}
}
\caption{\textbf{3-shot} evaluation performance (\%) on BBH.}
\label{tab:few-shot-detail}
\end{table}

%% file: tables/table_by_tasks.tex
\begin{table*}[t]
\small
\centering

\begin{tabular}{@{}lcccccccc@{}}
\toprule
& \multicolumn{2}{c}{\textbf{Human}\dag} & \multicolumn{2}{c}{\textbf{Llama-3-8B-}} &  \multicolumn{2}{c}{\textbf{Mistral-7B-}} &  \multicolumn{2}{c}{\textbf{CodeLlama-7B-}} \\
\textbf{Tasks} & \textbf{Avg.} & \textbf{Max} & \textbf{CoT} & \textbf{AoT} & \textbf{CoT} & \textbf{AoT} & \textbf{CoT} & \textbf{AoT} \\
\midrule
Boolean expressions* & 79.4 & 100.0 & 64.4 & 82.0 & 63.2 & 73.2 & 70.4 & 78.0 \\
Causal judgement & 69.6 & 100.0 & 56.7 & 61.0 & 56.7 & 63.6 & 52.4 & 57.2 \\
Date understanding & 76.8 & 100.0 & 53.6 & 16.4 & 47.6 & 40.8 & 37.6 & 21.2 \\
Disambiguation qa & 66.6 & 93.3 & 50.0 & 34.0 & 61.2 & 59.6 & 51.6 & 45.6 \\
Dyck languages* & 47.8 & 100.0 & 20.8 & 2.8 & 1.2 & 2.0 & 0.4 & 3.6 \\
Formal fallacies & 90.8 & 100.0 & 54.8 & 54.4 & 54.0 & 54.4 & 48.4 & 57.2 \\
Geometric shapes* & 54.0 & 100.0 & 20.0 & 26.0 & 30.8 & 21.6 & 20.0 & 34.0 \\
Hyperbaton & 74.7 & 100.0 & 49.2 & 71.6 & 72.4 & 49.2 & 69.6 & 62.8 \\
Logical deduction* & 40.3 & 88.9 & 50.3 & 52.5 & 45.1 & 51.9 & 36.7 & 37.2 \\
Movie recommendation & 60.7 & 90.0 & 60.4 & 62.0 & 60.0 & 63.6 & 48.0 & 58.0 \\
Multistep arithmetic two* & 9.7 & 25.0 & 4.4 & 98.0 & 1.2 & 100.0 & 0.8 & 96.8 \\
Navigate* & 81.9 & 100.0 & 61.6 & 62.8 & 50.4 & 64.8 & 45.6 & 57.6 \\
Object counting* & 86.1 & 100.0 & 50.0 & 66.0 & 45.6 & 85.2 & 42.4 & 46.8 \\
Penguins in a table & 78.0 & 100.0 & 54.1 & 52.7 & 44.5 & 50.7 & 45.2 & 51.4 \\
Reasoning about colored objects & 75.4 & 100.0 & 52.4 & 53.6 & 50.4 & 52.8 & 43.2 & 54.4 \\
Ruin names & 77.7 & 100.0 & 17.2 & 52.4 & 42.8 & 50.0 & 20.4 & 36.8 \\
Salient translation error detection & 36.7 & 80.0 & 40.4 & 46.4 & 44.8 & 51.2 & 39.6 & 43.6 \\
Snarks & 76.7 & 100.0 & 56.7 & 58.4 & 69.7 & 64.6 & 56.2 & 54.5 \\
Sports understanding & 70.8 & 100.0 & 70.8 & 57.2 & 42.0 & 59.6 & 52.0 & 54.8 \\
Temporal sequences* & 90.8 & 100.0 & 31.2 & 30.8 & 41.2 & 36.8 & 31.6 & 20.0 \\
Tracking shuffled objects* & 64.7 & 100.0 & 19.9 & 28.9 & 21.9 & 21.5 & 20.8 & 21.1 \\
Web of lies* & 81.3 & 100.0 & 52.0 & 57.6 & 51.6 & 54.4 & 49.6 & 51.6 \\
Word sorting* & 62.6 & 100.0 & 12.4 & 97.6 & 25.6 & 20.8 & 28.8 & 96.0 \\
\midrule
All Tasks & 67.7 & 94.4 & 43.6 & 53.3 & 44.5 & 51.8 & 39.6 & 49.6 \\
\bottomrule
\end{tabular}
\caption{Zero-shot performance (\%) on each task of BBH.
\dag indicates results from~\citet{DBLP:conf/acl/SuzgunSSGTCCLCZ23}.
* indicates that the task is an algorithmic task.}
\label{tab:bbh_by_tasks}
\end{table*}

%% file: tables/table_error.tex
\begin{table}[t]
\renewcommand{\arraystretch}{1.08} %
\small
\centering
\resizebox{\columnwidth}{!}{%
\begin{tabular}{@{}ll@{}}
\toprule
\textbf{Error Type (\%)} & \textbf{Description}  \\
\midrule
\multirow{2}{2.7cm}{Reasoning Error (38\%) }  &	\multirow{2}{5cm}{Errors due to deriving incorrect logical conclusions from given conditions.} 	\\
\\
\multirow{2}{2.7cm}{Knowledge Lacking (16\%)}  &	\multirow{2}{5cm}{Lack of world knowledge necessary to solve problems.} \\
\\
\multirow{2}{2.7cm}{Task Misunderstand- \\ ing (15\%)} &	\multirow{2}{5cm}{Failure to grasp the requirements or objectives of the task.}  \\
\\
\multirow{2}{2.7cm}{Condition Repetition (12\%)}  &	\multirow{2}{5cm}{Simply repeat the given conditions and then leap directly to the answer.}	 \\
\\
\multirow{2}{2.7cm}{Hallucination (9\%)} &	\multirow{2}{5cm}{Fabrication of non-existent conditions or misinterpretation of given conditions.}	\\
\\
\multirow{2}{2.7cm}{Runtime Error (5\%)} &	\multirow{2}{5cm}{Errors due to inexecutable programs or unformalized responses.}	 \\
\\
\multirow{2}{2.7cm}{False Negative (5\%)} &	\multirow{2}{5cm}{Correct answers are incorrectly identified as incorrect.}	 \\
\\
\bottomrule
\end{tabular}%
}
\caption{The failure modes of Llama-3-8B-AoT.}
\label{tab:table_error}
\end{table}

%% file: tables/table_aot_task_detail.tex
\onecolumn

{
\renewcommand{\arraystretch}{1.5}
\small

\begin{center}
\begin{longtable}{|p{0.2\textwidth}|p{0.1\textwidth}|p{0.6\textwidth}|}
\hline
\textbf{Dataset Name} & \textbf{Source} & \textbf{Task Category} \\ \hline
abductive nli & NIv2 & Story Composition, Coherence Classification \\ \hline
ade corpus v2 & NIv2 & Named Entity Recognition, Information Extraction, Text Categorization \\ \hline
adversarial qa@dbert & NIv2, T0 & Adversarial QA, Question Answering \\ \hline
adversarial qa@dbidaf & NIv2, T0 & Adversarial QA, Question Answering \\ \hline
adversarial qa@droberta & NIv2, T0 & Adversarial QA, Question Answering \\ \hline
afs & NIv2 & Text Matching \\ \hline
ai2 arc@ARC-Challenge & FLAN/T0 & Closed-Book QA, Inverted Closed-Book QA \\ \hline
ai2 arc@ARC-Easy & FLAN/T0 & Closed-Book QA, Inverted Closed-Book QA \\ \hline
ai2 arithmetic questions & NIv2 & Question Answering \\ \hline
air dialogue & NIv2 & Dialogue Generation, Speaker Identification, Intent Identification \\ \hline
ambigqa & NIv2 & Question Answering, Question Rewriting \\ \hline
anli r1 & FLAN/T0 & Natural Language Inference, Inverted Natural Language Inference \\ \hline
anli r2 & FLAN/T0 & Natural Language Inference, Inverted Natural Language Inference \\ \hline
anli r3 & FLAN/T0 & Natural Language Inference, Inverted Natural Language Inference \\ \hline
aqua & NIv2 & Question Answering \\ \hline
aquamuse & NIv2 & Answerability Classification \\ \hline
argkp & NIv2 & Text Matching \\ \hline
atomic & NIv2 & Commonsense Classification, Fill in The Blank \\ \hline
babi & NIv2 & Question Answering \\ \hline
bard & NIv2 & Word Analogy \\ \hline
big bench & NIv2 & Program Execution \\ \hline
bless & NIv2 & Word Relation Classification, Word Semantics \\ \hline
blimp & NIv2 & Linguistic Probing \\ \hline
break & NIv2 & Question Decomposition \\ \hline
cad & NIv2 & Toxic Language Detection \\ \hline
casehold & NIv2 & Text Completion \\ \hline
casino & NIv2 & Negotiation Strategy Detection \\ \hline
cb & NIv2, FLAN/T0 & Natural Language Inference, Inverted Natural Language Inference, Textual Entailment \\ \hline
cfq mcd1 & NIv2 & Text to Code \\ \hline
circa & NIv2 & Question Context Generation, Dialogue Generation, Text Matching \\ \hline
civil comments & NIv2 & Toxic Language Detection \\ \hline
clariq & NIv2 & Question Understanding, Dialogue Generation, Intent Identification \\ \hline
coached conv pref & NIv2 & Information Extraction, Speaker Identification \\ \hline
cod3s & NIv2 & Cause Effect Classification \\ \hline
coda 19 & NIv2 & Text Matching, Section Classification \\ \hline
codah & NIv2 & Text Completion \\ \hline
cola & NIv2, FLAN/T0 & Inverted Grammatical Acceptability, Text Quality Evaluation, Grammatical Acceptability, Grammar Error Detection \\ \hline
com2sense & NIv2 & Commonsense Classification \\ \hline
com qa & NIv2 & Question Answering, Question Rewriting \\ \hline
commonsenseqa & NIv2 & Question Answering \\ \hline
conala & NIv2 & Program Execution \\ \hline
conv ai 2 & NIv2 & Speaker Identification \\ \hline
cos e & T0 & Multiple-Choice QA (no trivia knowledge required) \\ \hline
cosmos qa & FLAN/T0 & Multiple-Choice QA (no trivia knowledge required), Inverted Multiple-Choice QA (no trivia knowledge required) \\ \hline
craigslist bargains & NIv2 & Text Categorization, Dialogue State Tracking \\ \hline
creak & NIv2 & Fact Verification \\ \hline
crows pair & NIv2 & Stereotype Detection \\ \hline
curated from stack overflow   english & NIv2 & Answerability Classification, Question Answering \\ \hline
curiosity dialogs & NIv2 & Information Extraction, Dialogue Generation, Speaker Identification \\ \hline
dailydialog & NIv2 & Text Categorization, Dialogue Act Recognition, Intent Identification, Sentiment Analysis \\ \hline
ddo & NIv2 & Text Categorization \\ \hline
deal or no dialogue & NIv2 & Dialogue State Tracking \\ \hline
defeasible nli atomic & NIv2 & Textual Entailment \\ \hline
detoxifying lms & NIv2 & Text Completion, Toxic Language Detection \\ \hline
dialogre & NIv2 & Speaker Relation Classification, Speaker Identification \\ \hline
diplomacy detection & NIv2 & Dialogue Generation \\ \hline
diqa & NIv2 & Answerability Classification, Question Answering \\ \hline
disfl qa & NIv2 & Answerability Classification, Question Answering, Text Quality Evaluation, Question Rewriting \\ \hline
dream & NIv2 & Question Understanding, Question Answering \\ \hline
drop & NIv2, FLAN/T0 & Inverted Mathematical QA, Mathematical QA, Question Understanding, Question Answering \\ \hline
dstc & NIv2 & Summarization, Dialogue State Tracking \\ \hline
duorc@ParaphraseRC & T0 & Extractive QA \\ \hline
duorc@SelfRC & T0 & Extractive QA \\ \hline
ecqa & CoT & Common Sense Reasoning Question Answering \\ \hline
emo & NIv2 & Sentiment Analysis \\ \hline
eqasc & NIv2 & Question Answering \\ \hline
equity evaluation corpus & NIv2 & Sentiment Analysis, Gender Classification, Fill in The Blank \\ \hline
esnli & CoT & Natural Language Inference \\ \hline
essential & NIv2 & Question Understanding, Answer Incomplete Questions \\ \hline
ethos & NIv2 & Toxic Language Detection \\ \hline
eurlex & NIv2 & Text Categorization \\ \hline
evaluation & NIv2 & Sentiment Analysis, Gender Classification, Fill in The Blank \\ \hline
event2mind & NIv2 & Emotion Word Generation, Sentiment Analysis, Emotional Reaction Generation, Intent Generation \\ \hline
financial phrasebank & NIv2 & Sentiment Analysis \\ \hline
freebase qa & NIv2 & Question Understanding, Question Answering \\ \hline
glucose & NIv2 & Cause Effect Classification, Information Extraction \\ \hline
go emotions & NIv2 & Sentiment Analysis \\ \hline
gooaq & NIv2 & Question Answering \\ \hline
google wellformed query & NIv2 & Question Understanding, Text Quality Evaluation \\ \hline
grailqa & NIv2 & Text Matching, Question Rewriting \\ \hline
gsm8k & CoT & Grade School Math Word Problems \\ \hline
gwsd & NIv2 & Textual Entailment \\ \hline
haspart kb & NIv2 & Entity Generation, Entity Relation Classification \\ \hline
hate speech offensive & NIv2 & Toxic Language Detection \\ \hline
hateeval & NIv2 & Toxic Language Detection \\ \hline
hatexplain & NIv2 & Toxic Language Detection \\ \hline
head qa & NIv2 & Text Categorization, Question Answering \\ \hline
health fact & NIv2 & Fact Verification, Sentence Composition, Explanation \\ \hline
help! need advice on identifying advice & NIv2 & Text Categorization \\ \hline
hippocorpus & NIv2 & Story Composition, Text Categorization \\ \hline
hope edi & NIv2 & Text Categorization \\ \hline
hybridqa & NIv2 & Question Answering, Pos Tagging \\ \hline
iirc & NIv2 & Answerability Classification, Question Answering \\ \hline
imppres & NIv2 & Textual Entailment, Sentence Composition \\ \hline
indian food 101 & NIv2 & Food Classification \\ \hline
inquistive & NIv2 & Question Answering \\ \hline
jeopardy & NIv2 & Answer Generation -- Jeopardy Difficulty Double, Answer Generation -- Jeopardy Difficulty Final, Answer Generation -- Jeopardy Difficulty Normal, Answer Generation -- Jeopardy Difficulty All \\ \hline
jigsaw & NIv2 & Toxic Language Detection \\ \hline
kilt tasks@hotpotqa & T0 & Closed-Book QA \\ \hline
leetcode & NIv2 & Text to Code \\ \hline
liar & NIv2 & Text Categorization, Keyword Tagging \\ \hline
math dataset & FLAN & Inverted Mathematical QA, Mathematical QA \\ \hline
math qa & NIv2 & Question Answering \\ \hline
mathmatics dataset & NIv2 & Question Understanding, Question Answering \\ \hline
mathqa & NIv2 & Question Answering \\ \hline
mcscript & NIv2 & Question Answering \\ \hline
mctaco & NIv2 & Answerability Classification, Text Quality Evaluation, Question Understanding, Question Answering \\ \hline
medical question pair dataset & NIv2 & Text Matching \\ \hline
meta woz & NIv2 & Dialogue Act Recognition \\ \hline
miam & NIv2 & Language Identification \\ \hline
miscellaneous & NIv2 & Question Understanding, Question Answering, Paraphrasing \\ \hline
missing & NIv2 & Question Answering \\ \hline
mnli & FLAN & Natural Language Inference, Inverted Natural Language Inference \\ \hline
mocah & NIv2 & Question Answering \\ \hline
mrqa & NIv2 & Question Answering \\ \hline
ms marco & NIv2 & Question Answering \\ \hline
multi woz v2 & NIv2 & Dialogue Generation, Speaker Identification \\ \hline
mutual & NIv2 & Dialogue Generation \\ \hline
mwsc & NIv2 & Question Answering \\ \hline
narrativeqa & NIv2 & Question Generation \\ \hline
natural questions & NIv2, FLAN/T0 & Closed-Book QA, Inverted Closed-Book QA, Question Answering \\ \hline
news headlines dataset for sacrasm detection & NIv2 & Text Categorization \\ \hline
nlg bias & NIv2 & Text Categorization \\ \hline
nlu asdiv dataset & NIv2 & Question Answering \\ \hline
numersense & NIv2 & Fill in The Blank \\ \hline
ohsumed & NIv2 & Information Extraction, Text Matching, Keyword Tagging \\ \hline
olid & NIv2 & Toxic Language Detection \\ \hline
open pi & NIv2 & Text Categorization \\ \hline
openbookqa & NIv2, FLAN/T0 & Multiple-Choice QA (no trivia knowledge required), Sentence Composition, Inverted Multiple-Choice QA (no trivia knowledge required), Question Answering \\ \hline
opp 115 & NIv2 & Information Extraction, Text Categorization \\ \hline
overruling & NIv2 & Text Categorization \\ \hline
persent & NIv2 & Named Entity Recognition, Sentiment Analysis \\ \hline
personachat & NIv2 & Dialogue Generation \\ \hline
perspectrum & NIv2 & Textual Entailment \\ \hline
piqa & NIv2, FLAN/T0 & Multiple-Choice QA (no trivia knowledge required), Inverted Multiple-Choice QA (no trivia knowledge required), Question Answering \\ \hline
poki & NIv2 & Poem Generation, Text Categorization \\ \hline
propara & NIv2 & Named Entity Recognition, Information Extraction \\ \hline
prost & NIv2 & Question Generation \\ \hline
proto qa & NIv2 & Question Answering \\ \hline
pubmed qa & NIv2 & Answer Verification, Intent Identification, Question Answering \\ \hline
qa srl & NIv2 & Question Answering \\ \hline
qanta & NIv2 & Text Categorization \\ \hline
qasper & NIv2 & Question Understanding, Question Answering \\ \hline
qed & NIv2 & Question Answering \\ \hline
qnli & FLAN/T0 & Multiple-Choice QA (no trivia knowledge required), Inverted Multiple-Choice QA (no trivia knowledge required) \\ \hline
qrecc & Dialog & Conversational Question Answering \\ \hline
qrecc input inversion & Dialog & Conversational Question Answering \\ \hline
quac & FLAN/T0 & Multiple-Choice QA (no trivia knowledge required), Inverted Multiple-Choice QA (no trivia knowledge required) \\ \hline
quail & NIv2 & Question Answering \\ \hline
quarel & NIv2 & Question Answering \\ \hline
quartz & NIv2 & Explanation, Question Answering \\ \hline
question \& answer zre & NIv2 & Question Understanding, Question Answering \\ \hline
quoref & NIv2 & Question Answering \\ \hline
race@high & T0 & Multiple-Choice QA (no trivia knowledge required), Span Generation \\ \hline
race@middle & T0 & Multiple-Choice QA (no trivia knowledge required), Span Generation \\ \hline
recepie nlg & NIv2 & Named Entity Recognition, Fill in The Blank \\ \hline
roc stories & NIv2 & Text Completion, Sentence Ordering, Coherence Classification \\ \hline
root09 & NIv2 & Word Relation Classification, Word Semantics \\ \hline
ropes & NIv2 & Story Composition, Question Answering \\ \hline
rte & NIv2, FLAN/T0 & Natural Language Inference, Inverted Natural Language Inference, Textual Entailment \\ \hline
ruletaker & NIv2 & Fact Verification \\ \hline
sarcasm in twitter & NIv2 & Text Categorization \\ \hline
sbic & NIv2 & Toxic Language Detection \\ \hline
schema guided dstc8 & NIv2 & Text Categorization, Dialogue Act Recognition \\ \hline
scifact & NIv2 & Text Matching, Text Quality Evaluation \\ \hline
sciq & NIv2 & Explanation, Question Answering \\ \hline
scitail & NIv2 & Textual Entailment, Sentence Composition, Question Answering \\ \hline
scitailv1.1 & NIv2 & Textual Entailment, Sentence Composition \\ \hline
scitldr & NIv2 & Summarization \\ \hline
scruples & NIv2 & Text Categorization, Ethics Classification \\ \hline
semeval 2018 task3 & NIv2 & Irony Detection \\ \hline
semeval 2019 task 10 & NIv2 & Question Answering \\ \hline
semeval 2020 task4 & NIv2 & Commonsense Classification, Explanation \\ \hline
semeval 2020 task 7 & NIv2 & Text Categorization \\ \hline
smcalflow & NIv2 & Dialogue Generation, Speaker Identification \\ \hline
snli & NIv2, FLAN & Natural Language Inference, Sentence Composition, Inverted Natural Language Inference, Textual Entailment \\ \hline
social i qa & T0 & Multiple-Choice QA (no trivia knowledge required) \\ \hline
spolin & NIv2 & Dialogue Generation, Dialogue Act Recognition \\ \hline
squad v1 & FLAN/T0 & Extractive QA, Inverted Extractive QA \\ \hline
squad v2 & FLAN/T0 & Extractive QA, Inverted Extractive QA \\ \hline
starcon & NIv2 & Text Matching, Stance Detection \\ \hline
stereoset & NIv2 & Text Categorization, Stereotype Detection, Fill in The Blank \\ \hline
storycommonsense & NIv2 & Information Extraction, Intent Identification, Sentiment Analysis \\ \hline
strategyqa & NIv2 & Question Decomposition \\ \hline
subjqa & NIv2 & Question Answering \\ \hline
super glue@boolq & NIv2 & Question Answering \\ \hline
super glue@multirc & NIv2, FLAN/T0 & Answerability Classification, Question Answering, Inverted Multiple-Choice QA (no trivia knowledge required), Multiple-Choice QA (no trivia knowledge required), Answer Verification, Text Quality Evaluation \\ \hline
super glue@record & NIv2, FLAN/T0 & Question Answering, Extractive QA, Inverted Extractive QA \\ \hline
svamp & NIv2 & Question Answering \\ \hline
swag & NIv2 & Text Completion \\ \hline
synthetic & NIv2 & Clock Format Conversion, Information Extraction, Program Execution, Spelling Error Detection, Grammar Error Detection, Rhyme Generation, Pos Tagging, Date Validity Prediction, Mathematics, Temporal Reasoning, Leap Year Prediction, Edible Prediction \\ \hline
task master & Dialog & Dialog Next Turn Prediction \\ \hline
task master input inversion & Dialog & Dialog Next Turn Prediction \\ \hline
tellmewhy & NIv2 & Answerability Classification, Question Answering \\ \hline
timetravel & NIv2 & Story Composition, Coherence Classification \\ \hline
tom qa & NIv2 & Question Answering \\ \hline
torque & NIv2 & Information Extraction, Question Answering \\ \hline
trianglecopa & NIv2 & Question Generation \\ \hline
trivia qa & FLAN/T0 & Extractive QA, Inverted Extractive QA \\ \hline
tweetqa & NIv2 & Answerability Classification, Answer Verification, Question Answering \\ \hline
web questions & NIv2, T0 & Closed-Book QA, Question Answering \\ \hline
wiki dialog & Dialog & Conversational Question Answering \\ \hline
wiki dialog input inversion & Dialog & Conversational Question Answering \\ \hline
wiki hop & NIv2, T0 & Multiple-Choice QA (no trivia knowledge required), Question Answering \\ \hline
wiki movies & NIv2 & Question Answering \\ \hline
wiki qa & NIv2, T0 & Answer Verification, Closed-Book QA \\ \hline
wikitext 103 & NIv2 & Fill in The Blank \\ \hline
winowhy & NIv2 & Coreference Resolution, Commonsense Classification, Explanation \\ \hline
wiqa & NIv2 & Sentence Ordering, Question Answering \\ \hline
wnli & FLAN & Coreference Resolution, Inverted Coreference Resolution \\ \hline
x\_csr & NIv2 & Question Answering, Linguistic Probing \\ \hline
xcsr & NIv2 & Sentence Perturbation \\ \hline
xl wic & NIv2 & Word Semantics, Sentence Composition \\ \hline
xquad & NIv2 & Question Answering \\ \hline
yahoo answers topics & NIv2 & Text Categorization \\ \hline
yoruba bbc topics & NIv2 & Text Categorization \\ \hline

\caption{The datasets included in our \ourdataset, along with their sources and task categories. The sources and task categories are aligned with the FLAN Collection (\url{https://github.com/google-research/FLAN/blob/main/flan/v2/flan_collection_info.csv}).
}
\label{tab:aot_task_detail}
\end{longtable}
\end{center}

}

\twocolumn